\documentclass[manuscript, screen]{acmart}

\AtBeginDocument{%
  }

\setcopyright{acmlicensed}
\copyrightyear{2025}
\acmYear{2025}
\acmDOI{XXXXXXX.XXXXXXX}
\acmConference[Conference acronym 'XX]{Make sure to enter the correct
  conference title from your rights confirmation email}{June 03--05,
  2018}{Woodstock, NY}
\acmISBN{978-1-4503-XXXX-X/2018/06}

\usepackage{svg}
\usepackage{xcolor}
\usepackage{soul}
\usepackage{multirow}
\usepackage{graphicx}
\usepackage{pifont}
\usepackage{subcaption}
\usepackage{xcolor}

\newcommand{\checkmarks}{%
\ding{51}
}

\newcommand{\xmark}{%
\ding{55}
}

\newif\ifshowchanges
\showchangesfalse
\newcommand{\changed}[1]{
  \ifshowchanges            
    {\color{red}#1}%
  \else 
    #1%
  \fi
}

\begin{document}

\title[Event-based audio processing on SoC FPGA]{Hardware-accelerated graph neural networks: an alternative approach for
event-based audio classification and keyword spotting on SoC FPGA}


\author{Kamil Jeziorek}
\email{kjeziorek@agh.edu.pl}
\orcid{0000-0001-5446-3682}
\author{Piotr Wzorek}
\email{pwzorek@agh.edu.pl}
\orcid{0000-0003-3885-600X}
\affiliation{%
  \institution{The AGH University of Krakow}
  \city{Krakow}
  \country{Poland}}

\author{Krzysztof Błachut}
\email{kblachut@agh.edu.pl}
\orcid{0000-0002-1071-335X}
\affiliation{%
  \institution{The AGH University of Krakow}
  \city{Krakow}
  \country{Poland}}

\author{Hiroshi Nakano}
\orcid{0009-0009-5493-4689}
\email{nakano@west.sd.keio.ac.jp}
\affiliation{%
  \institution{Graduate School of Science and Technology, Keio University}
  \city{Tokyo}
  \country{Japan}}

\author{Manon Dampfhoffer}
\orcid{0000-0002-2622-3101}
\email{Manon.DAMPFHOFFER@cea.fr}
\author{Thomas Mesquida}
\orcid{0000-0002-2572-5353}
\email{thomas.mesquida@cea.fr}
\affiliation{%
  \institution{CEA-List, Université Grenoble Alpes}
  \city{Grenoble}
  \country{France}}

\author{Hiroaki Nishi}
\orcid{0000-0002-6331-2947}
\email{west@keio.jp}
\affiliation{%
  \institution{Graduate School of Science and Technology, Keio University}
  \city{Tokyo}
  \country{Japan}}

\author{Thomas Dalgaty}
\orcid{0000-0003-0326-2121}
\email{Thomas.DALGATY@cea.fr}
\affiliation{%
  \institution{CEA-List, Université Grenoble Alpes}
  \city{Grenoble}
  \country{France}}

\author{Tomasz Kryjak}
\orcid{0000-0001-6798-4444}
\email{kryjak@agh.edu.pl}
\affiliation{%
  \institution{The AGH University of Krakow}
  \city{Krakow}
  \country{Poland}}

\renewcommand{\shortauthors}{Jeziorek et al.}

\begin{abstract}

As the volume of data recorded by embedded edge sensors increases, particularly from neuromorphic devices producing discrete event streams, there is a~growing need for hardware-aware neural architectures that enable efficient, low-latency, and energy-conscious local processing.
We present an FPGA implementation of event-graph neural networks for audio processing. We utilise an artificial cochlea that converts time-series signals into sparse event data, reducing memory and computation costs. Our architecture was implemented on a~SoC FPGA and evaluated on two open-source datasets. For classification task, our baseline floating-point model achieves 92.7\% accuracy on SHD dataset -- only 2.4\% below the state of the art -- while requiring over 10$\times$ and 67$\times$ fewer parameters. On SSC, our models achieve 66.9–71.0\% accuracy. Compared to FPGA-based spiking neural networks, our quantised model reaches 92.3\% accuracy, outperforming them by up to 19.3\% while reducing resource usage and latency. For SSC, we report the first hardware-accelerated evaluation.
We further demonstrate the first end-to-end FPGA implementation of event-audio keyword spotting, combining graph convolutional layers with recurrent sequence modelling. The system achieves up to 95\% word-end detection accuracy, with only 10.53 $\mu$s latency and 1.18 W power consumption, establishing a~strong benchmark for energy-efficient event-driven KWS.
\end{abstract}




\maketitle

\section{Introduction}

\begin{figure}[!t]
    \centering
    \includegraphics[width=0.99\linewidth]{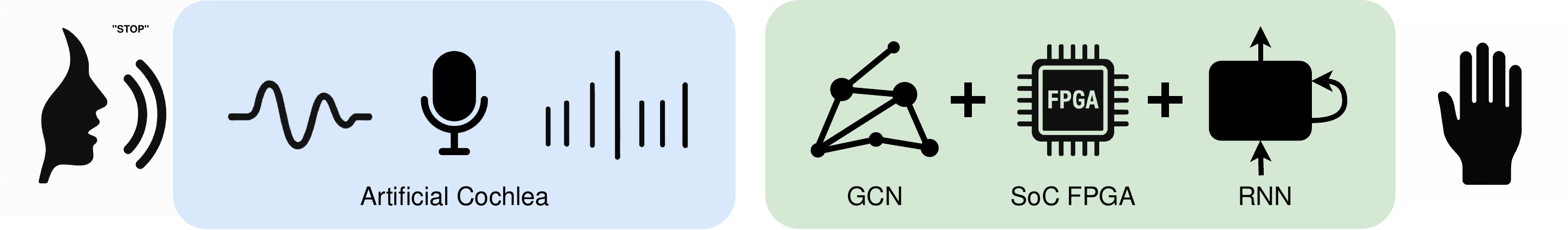}
    \caption{In this work, we propose an event-based keyword spotting system in which speech signals are converted into asynchronous events by an artificial cochlea and represented as spectro-temporal event-graphs. These are processed by a~GCN–RNN model deployed on a~SoC FPGA, enabling low-power, low-latency, and efficient keyword spotting.}
    \label{fig:teaser}
    \Description{The figure illustrates an event-based keyword spotting system. Speech signals are converted into asynchronous events by an artificial cochlea and represented as spectro-temporal event-graphs. These are processed by a~GCN–RNN model deployed on a~SoC FPGA, enabling low-power, low-latency, and efficient keyword detection.}
\end{figure}

As the Internet of Things expands, distributed sensors are collecting ever-increasing quantities of data. 
This has driven the need for accurate and efficient computing systems capable of processing this data locally, for example, to make predictions \cite{ren2023deep}. 
The energy consumption and latency of these systems are of particular importance, as data is increasingly processed directly on edge devices, which operate under strict energy constraints (e.g. the battery of a~smartwatch).

In most cases, the raw data produced by the sensors is time-series -- continuous signals following the evolution of environmental variables \cite{lim2021time}. For example, sensors that monitor the vibration of mechanical parts have been used to predict failures in gearboxes \cite{saufi2020gearbox}, or implantable cardioverter-defibrillators monitor the state of a~patient's heart in order to apply an electric shock in the event of dangerous fibrillation \cite{dimarco2003implantable}. 

It is becoming increasingly common to use artificial intelligence (AI) methods to process this time-series data. However, using conventional hardware, such as GPUs (graphics processing units), consumes too much power, and microprocessors may struggle to meet the latency requirements of many applications. 
FPGAs (Field Programmable Gate Arrays) and custom integrated circuits offer a~means of implementing architectures optimised to specific AI methods that are capable of meeting latency requirements while minimising power consumption \cite{Al-Ameri,Al-Ali,Guo}.
Furthermore, it is necessary to apply the model at periodic intervals, and important temporal information present in the signal below the sampling frequency cannot be leveraged.

A~particularly promising method is event-based AI, which operates on the sparse data generated by neuromorphic, event-based sensors and allows reducing power consumption and prediction latency \cite{gallego2020event}. Event-based sensors, instead of regularly sampling an environmental variable, generate ``events'' in case of the changes in the signal. 

In this work, which is an extended version of our conference paper \cite{nakano2025arc} presented at the \textit{21st International Symposium on Applied Reconfigurable Computing}, we focus on audio signal processing.
In this context, event-based time-series data is generated by a~class of sensors known as artificial cochleas (AC) (also referred to as dynamic audio sensors  or silicon cochleas) \cite{liu2013asynchronous,mostafa2024}.
Their operating principle is to apply a~bank of band-pass filters to separate the signal into multiple frequency channels. A~digital pulse (i.e. an event) is generated per channel in an asynchronous manner when the signal intensity changes by a~pre-defined threshold. This results in a~sparse spectrogram, 
which an event-based AI method exploits to perform efficient computation. 

Processing AC-generated event-data has been the subject of many publications. 
The most common approach is to apply spiking neural networks (SNNs) implemented for specialised hardware \cite{ortner2023online,carpegna2024spiker,dalgaty2024mosaic}. However, it is not clear how event-data sparsity can be truly exploited due to the nondeterministic pattern of synaptic weight-memory access \cite{dampfhoffer_tetci_2022,dalgaty2023cnn} inherent to SNNs. 

Recently, event-graph neural networks have been proposed as an alternative way of processing event-data \cite{li2021graph,dalgaty2023hugnet,mesquida2023g2n2,jeziorek2023memory,lars}.
The event-graph approach consists of a~dynamically updated graph generated by an event-sensor, and it involves applying graph convolutions on the resulting data structure.
Unlike SNNs, the weight access pattern for many event-graph models is deterministic. This may provide an opportunity to develop new event-based AI hardware that is truly capable of exploiting the inherent sparsity of data to reduce power consumption and latency. Although digital architectures for accelerating event-graphs have been proposed in the context of computer vision \cite{jeziorek2024embedded,yang2024evgnn}, a~dedicated architecture for time-series audio applications has not yet been considered. 

In this paper, we propose a~hardware accelerator implemented on a~SoC FPGA for event-graph audio classification and keyword spotting tasks (Figure~\ref{fig:overview}). The proposed method enables real-time, end-to-end continuous processing while preserving the inherent sparsity of the input data.
Specifically, we consider a~recently proposed spectro-temporal model \cite{lars} developed for the classification of time-series data and evaluated on the Spiking Heidelberg Digits (SHD) dataset \cite{cramer2020}, which is a~representative of time-series data.


We summarise our main contribution as follows:

\begin{itemize}
\item  We use the hardware-aware design method to propose optimisations required to implement spectro-temporal event-graphs in reconfigurable hardware with low power, low latency and low resource utilisation. \changed{These optimisations include modifications to the graph generation module, the introduction of additional normalisation in graph convolution, and the careful selection of model hyperparameters.}

\item  We propose the first embedded system for event-graph-based audio processing on a~SoC FPGA and also the first hardware implementation that supports fully asynchronous event-by-event processing with conservation of data's temporal sparsity.

\item We achieve a~new state-of-the-art performance for time-series data classification on an FPGA applied to the Spiking Heidelberg Digits dataset, demonstrating significant improvements in resource utilisation, latency reduction and accuracy compared to previous SNN-based approaches. \changed{Moreover, we introduce the first FPGA-based benchmark results on the SSC dataset, establishing a~reference point for future research.}

\item  \changed{We propose the first end-to-end hardware implementation of the keyword spotting task model using event-based audio, designed for continuous and real-time processing. The architecture employs recurrent layers and achieves high word-end detection accuracy while maintaining low latency and efficient resource utilisation.}

\end{itemize}

The remainder of this paper is organised as follows.
In Section \ref{sec:related_work} we present an overview of the related work.
In Section \ref{sec:proposed_system} we introduce the proposed modification and embedded audio processing system implementation.
In Section \ref{sec:evaluation} we present the results of ablation studies and comparisons with the state-of-the-art.
We conclude with a~discussion on future research directions and summary in Section \ref{sec:summary}.

\section{\changed{Related work}}
\label{sec:related_work}

\subsection{Audio processing}

In the domains of IoT devices and mobile robotics, there is a~growing demand for efficient audio signal processing, particularly for the speech.
One of the main tasks in this context is automatic speech recognition (ASR), which converts the entire speech segment into text.
There are several techniques and datasets used for this task \cite{ahlawat2025survey}.

A~special case of the ASR is keyword spotting (KWS), in which only specific, predefined words are recognised from a~continuous speech.
This makes KWS applications much smaller and computationally effective, as not all words need to be recognised.
Therefore, they are increasingly used in embedded devices, microcontrollers, as e.g. voice assistants.

For many years, the most popular approach for this task was to use so-called mel-frequency cepstral coefficients (MFCC) \cite{davis1980} along with hidden Markov models (HMM) \cite{rohlicek1989}.
However, huge popularity of deep neural network-based approaches in various tasks motivated researchers to apply them in the audio domain as well, with the first work \cite{chen2014} published in 2014.
Since then, NN-based solutions are much more widely used.
For thorough overview of different KWS methods check an excellent paper \cite{lopez2021kwsoverview}.

The most popular dataset used for KWS development and evaluation is the publicly available Google Speech Command Dataset (GSCD) \cite{warden2018googledataset}.
Its first version was released in 2017, while the second one year later, in which additional words and more speakers were added to extend the dataset.
Currently, there are 35 words, over 100k speech segments and over 2600 different speakers.
The recordings were made by smartphone and laptop microphones, they are therefore quite noisy.
Currently, top solutions achieve over 98\% accuracy on this dataset \cite{kim21l_interspeech}.

In this work, we distinguish between two related tasks: \emph{word classification}, which identifies a~given word based on a~single audio sample without specifying its temporal occurrence, and \emph{keyword spotting}, which detects and classifies a~given word immediately after its occurrence, with the possibility of selecting a~predefined set of keywords.

\subsection{Event-audio data}
\label{sec:event-audio-related}

Contrary to conventional microphones, which produce uniformly sampled waveforms, the artificial cochlea emits asynchronous events that capture temporal contrast in the acoustic scene.
The invention of AC models, especially dynamic audio sensors (DAS) \cite{liu2013asynchronous,mostafa2024}, which are inspired by biological auditory systems, emerged an opportunity to apply event-based AI methods, extensively used in computer vision, for time-series applications. 

The most widely used benchmark for evaluating event-based time-series models is the Spiking Heidelberg Digits (SHD) dataset \cite{cramer2020}. It simulates an AC by filtering recordings with a~computational model of the inner ear. The dataset comprises more than 10,000 recordings (8156 for training and 2264 for testing), consisting of spoken digits from zero to nine in both English and German. Each recording yields a sparse 700-channel spectrogram, with an average duration of 750 ms (Figure \ref{fig:events_hist}).
In order to compare our SoC FPGA architecture of an event-graph neural network to previous state-of-the-art software and hardware implementations, we use the SHD dataset with the same train/test split.

The second benchmark employed is the Spiking Speech Command (SSC) dataset, derived from the Google Speech Commands Dataset (GSCD) by converting the recordings into spike-based representations \cite{cramer2020}, simulating the output of an event-based sensor, specifically an artificial cochlea. The dataset includes 75,466 training samples, 9,981 validation samples, and 20,382 test samples, covering 35 word categories spoken by a~diverse set of speakers.

Existing works evaluated on both datasets
mostly contain software implementations of recurrent spiking neural networks \cite{cramer2020,bittar2022,dampfhoffer2022,rossbroich2022,ceolini2019} and feed-forward models leveraging learned synaptic delays \cite{dagostino2024,hammouamri2023,malettira2024,sun2023,yu2022}.
There are also articles \cite{blouw2020,pedroni2018} devoted to comparison between DNNs and SNNs for keyword spotting applications with the latter around 4-5$\times$ more energy-efficient.
The article \cite{dominguez2018nas} presents the usage of a~neuromorphic auditory sensor (NAS), which is a~digital version of artificial cochlea implemented in FPGA.
The authors used its output to train and evaluate a~spiking convolutional neural network on custom data.
A~very similar approach was taken by the authors of the work \cite{xu2023event}, in which the NAS sensor was also realised in FPGA and its output was used to train and evaluate an SNN on custom-generated events from the TIDIGITS dataset.

More recently, an alternative approach was proposed in \cite{lars}, based on a~fundamentally different paradigm. Instead of directly applying an SNN to the event stream, the data is first transformed into a~graph representation and subsequently processed by a~graph neural network (GNN). This event-graph neural network achieved performance comparable to state-of-the-art synaptic delay-based SNNs on the SHD benchmark, while requiring one to two orders of magnitude fewer synaptic weights in many cases. Notably, a~compact model with only 17k parameters achieved a~test accuracy exceeding $90\%$. These findings demonstrate the potential of event-graph methods for processing event-based time-series data. While it is believed that event-graphs, due to deterministic synaptic weight access patterns and natively asynchronous operation, may translate well into a~dedicated hardware, it needs to be thoroughly investigated.

\subsection{Event-audio processing on FPGA}
\label{sec:fpga-related}

Driven by the demand for real-time and low-power systems, numerous dedicated hardware implementations of event-based processing algorithms have been proposed on both application-specific integrated circuits \cite{basu2022spiking} as well as FPGAs \cite{kryjak2024event}. The majority of these works have focused on event-vision applications. 
To the best of the authors' knowledge, there are only two research papers describing FPGA-based SNN implementations for AC time-series processing \cite{carpegna2024spiker,quantisenc2024}.
Both of them report benchmarking results on the SHD dataset and consider only the simple classification for samples containing single words.


The Spiker+ framework \cite{carpegna2024spiker} and QUANTISENC tool \cite{quantisenc2024} both target efficient SNN accelerators for FPGA-based edge computing, with Spiker+ achieving 72.9\% accuracy on the SHD benchmark at 430 mW and 540 $\mu$s latency, while QUANTISENC demonstrated 87.8\% accuracy and 1.6 W peak power consumption. 


This work is an extension of \cite{nakano2025arc}, where, drawing inspiration from the computer vision domain \cite{jeziorek2024embedded,yang2024evgnn}, we introduced an event-graph neural network implemented on the Xilinx ZCU104 SoC FPGA for SHD audio classification with low latency and low power requirements.
By leveraging a~graph-based representation, we were able to drastically reduce the number of model parameters compared with state-of-the-art solutions, with only a~marginal accuracy decrease (approximately 2\%), thereby establishing new state-of-the-art results for FPGA-based event-audio processing.

Beyond event-based implementations, several works have explored FPGA accelerators for keyword spotting using conventional audio waveforms.
In \cite{yoon2023}, a~binary convolutional neural network (BCNN) was implemented on an Intel Cyclone V board, achieving 91.6\% accuracy on the GSCD dataset. Similarly, a~convolutional neural network (CNN) was deployed in \cite{mourrane2023} on a~DE2-115 board with an Intel FPGA, reaching 90.4\% accuracy on GSCD. In \cite{krishna2023}, a~custom NN-based accelerator, RAMAN, was proposed for audio classification on the Microchip MPFS250T SoC FPGA, although no evaluation metrics were reported. Another study \cite{ng2024} implemented a~separable CNN (SCNN) on a~Digilent Arty A7-35T FPGA board, achieving over 90\% accuracy on the MLPerf Tiny benchmark, though without reporting results on other datasets.

Given the promising potential of the event-graph approach, in this work we propose an end-to-end system based on graph convolutional neural networks for classification and keyword spotting tasks. We compare accuracy, power consumption, latency, and resource utilisation against prior FPGA-based implementations on publicly available benchmarks. This analysis enables us to assess whether the performance advantages of event-graph neural networks translate effectively to hardware for continuous real-time processing, and how they compare to SNN-based designs.

\section{The proposed method}
\label{sec:proposed_system}

\begin{figure}[t]
    \centering
    \includegraphics[width=0.99\linewidth]{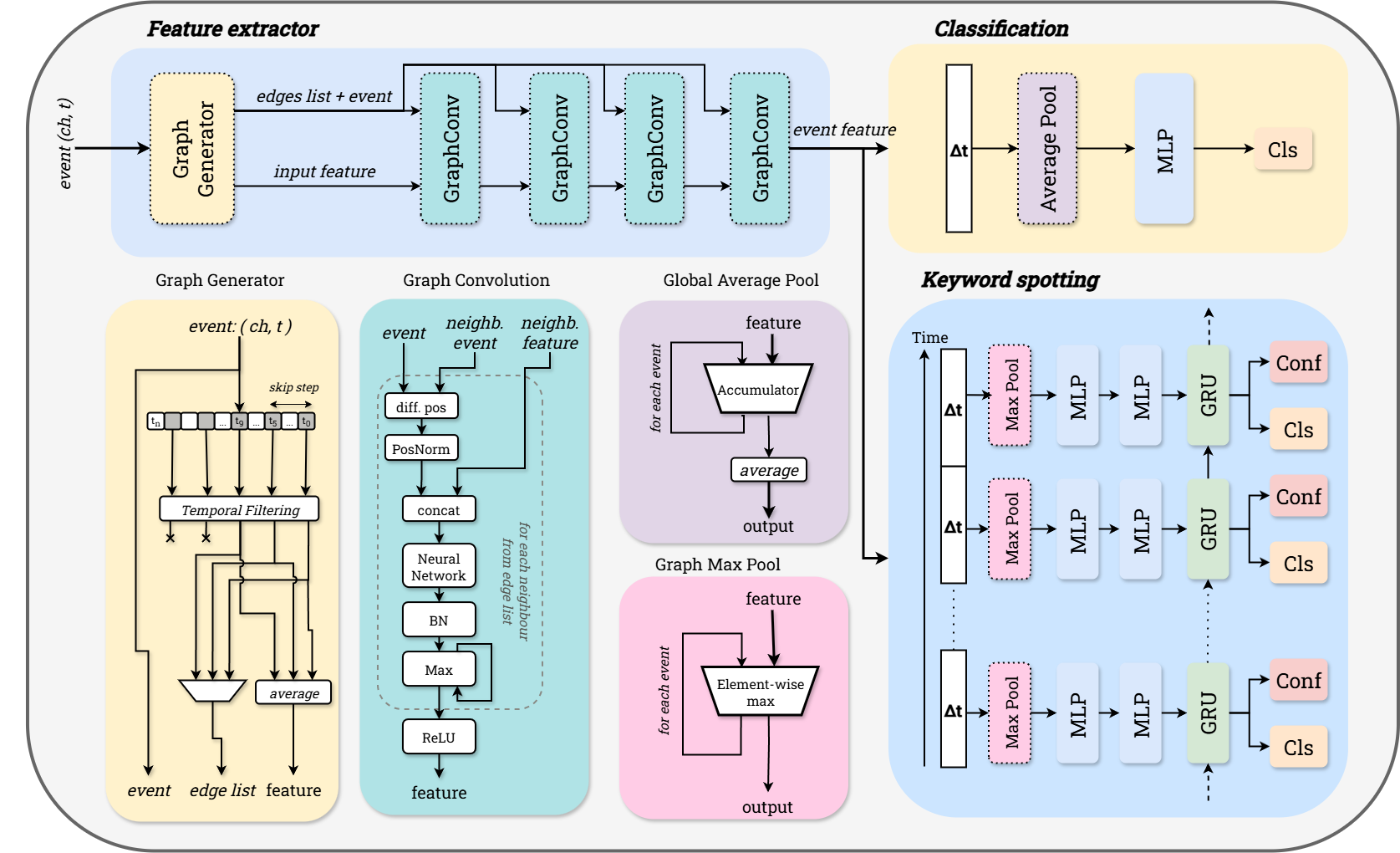}
    \caption{\textbf{Overview of our hardware-accelerated event-graph neural network implementation.} Events from the artificial cochlea are first read and preprocessed in the processing system (PS), and subsequently processed by an asynchronous graph neural network in the programmable logic (PL). Modules marked with dotted lines operate on an event-by-event basis. For the classification task, we employ a~global average pooling layer, which updates an accumulator over the entire sample and returns the result to a~fully connected classifier (MLP). For keyword spotting, the data is divided into smaller time windows ($\Delta$t), aggregated using a~graph max pooling operation, and processed sequentially by two MLPs and a~GRU, after which confidence and class scores are computed. In both tasks, we employ the same feature extractor, consisting of a~graph generator followed by four graph convolution layers.}
    
    \Description{The figure presents the hardware-accelerated event-graph neural network implementation. Events from the artificial cochlea are converted into event-graphs, processed by graph convolution layers, and shared between two tasks: classification and keyword spotting. Classification uses global average pooling followed by an MLP, while keyword spotting divides data into time windows, applies graph max pooling, and processes it with MLPs and a~GRU to compute confidence and class scores.}
    
    \label{fig:overview}
\end{figure}

\changed{In this section we introduce our method for low-latency classification and keyword spotting that couple an artificial cochlea with an FPGA. 
We exploit the sparsity of this neuromorphic sensor with a~hardware-aware design of a~graph convolutional network as a~streaming feature extractor operating on event-derived graphs. 
The design minimises end-to-end latency and sustains constant throughput under bursty event rates.
To generate a~final prediction, we apply pooling across event embeddings from the last convolutional layer. 
The resulting vector is then fed into a~network's head, which produces the desired output for the processed event data.

As shown in Figure \ref{fig:overview}, classification and keyword spotting architectures share the same feature-extraction module but employ task-specific heads (which we describe in detail later).}

\subsection{Feature extraction}
\label{sec:featureextraction}

The principle of the event-graph approach is to generate a~graph from raw events generated by an event-sensor -- in this case the artificial cochlea.
Formally, a~graph is defined as $\mathcal{G} = (\mathcal{V},\mathcal{E})$, where $\mathcal{V}$ is the set of vertices and $\mathcal{E}$ is the set of edges. Each vertex $v \in \mathcal{V}$ is associated with corresponding event's position $\mathcal{P}$ and feature vector $\mathcal{X}$, both of which characterise the underlying entity that the event represents, while edges represent the connections between pairs of these entities.

Spectro-temporal event-graphs are a~specific form of event-graphs constructed from time-series data. They are typically created by performing a~hemispherical search in the channel-time domain (Figure \ref{fig:events_hist}), establishing edges between events that lie within defined distance.
This neighbourhood can be determined by spatial and temporal radius thresholds, or, alternatively, edges can be assigned randomly within the search volume. Each new event generated by the sensor forms directed edges from previously recorded events found within a~semi-circle defined by a~channel radius $r_{ch}$ and a~time radius $r_t$. 
\changed{By employing this method, the graph can be updated for each new event and resulting vertex can be processed directly, eliminating the need for data aggregation within predefined time windows. This significantly reduces the overall system latency.}

Given an event-graph, we can apply PointNetConv \cite{POINTNET} operations across \(L\) convolutional layers. Each layer uses a~unique weight matrix shared across all events. Within each layer, the following operation is performed to update the embedding of the \(i^{th}\) event:

\begin{equation}
\label{eq:pointnetconv}
\hat{\mathcal{X}}_i = \max_{j \in \mathcal{N}(i)} \bigl( \phi([\mathcal{X}_j \;||\; (\mathcal{P}_j - \mathcal{P}_i)]) \bigr),
\end{equation}

\noindent where \(\hat{\mathcal{X}}_i\) is the updated feature vector for event \(i\), and \(\phi(\cdot)\) represents a~fully-connected layer. The notation \(||\) indicates the concatenation of the feature vector \(\mathcal{X}_j\) with the relative position vector \(\mathcal{P}_j - \mathcal{P}_i\). After this transformation, a~feature-wise max pooling aggregates the neighbour contributions into a~single output vector per event, and a~ReLU activation is applied to introduce nonlinearity.

The following subsections detail the adaptations and optimisations performed in order to map feature extraction baseline \cite{lars} to a~hardware accelerator architecture on FPGA.

\subsubsection{\textbf{Graph generator}}
\label{sec:graphgenerator}

The baseline model presented in \cite{lars} has proven effective in capturing spectro-temporal relationships within event-graphs. However, significant challenges are faced when considering hardware implementation. 

In \cite{lars}, the feature vector \(\mathcal{X}\) of each event consisted of two components of a~normal vector estimated by fitting a~local surface to the event-data using a~least-squares approach. 
The position \(\mathcal{P}\) comprised a~channel index and a~timestamp. The relative positions between events defined edge vectors, creating an \(\mathcal{N}(i)\) neighbourhood set for each node \(i\).

However, the calculation of a~normal vector requires fitting regression lines to the event-data that has a~high computational complexity. 
Moreover, identifying neighbouring vertices requires storing all vertex vectors in memory and performing sequential searches, resulting in high latency and significant memory overhead.

To address these limitations, a~novel graph generation method was designed to optimise memory usage, reduce latency, and minimise computational overhead. The following key modifications were introduced:

\begin{enumerate}
    \item Drawing inspiration from FPGA-based event camera data processing implementations \cite{jeziorek2024embedded,yang2024evgnn}, events are stored in 1D context memory (implemented as a~block RAM memory (BRAM)) using their channel \( ch \) as the address and timestamps \( t \) as the data. Only the most recent event generated per channel is stored, continuously overwriting timestamps at each channel index. Each new event can be connected only to the ones already processed, creating a~directed graph. \changed{The number of neighbours with the same channel $ch$ is thus limited to 1, significantly reducing the required memory usage}.
    
    \item Event-normal vectors were replaced with simpler features based on the average timestamp and channel coordinates of neighbouring events, massively reducing the computational cost.

    \item To improve the efficiency of neighbour search, we introduced a~method called \textit{skip step connection}. This corresponds to a~pre-defined deterministic pattern regarding how edges can be formed between vertices.
\end{enumerate}

\begin{figure}[t]
    \centering
\includegraphics[width=0.8\linewidth]{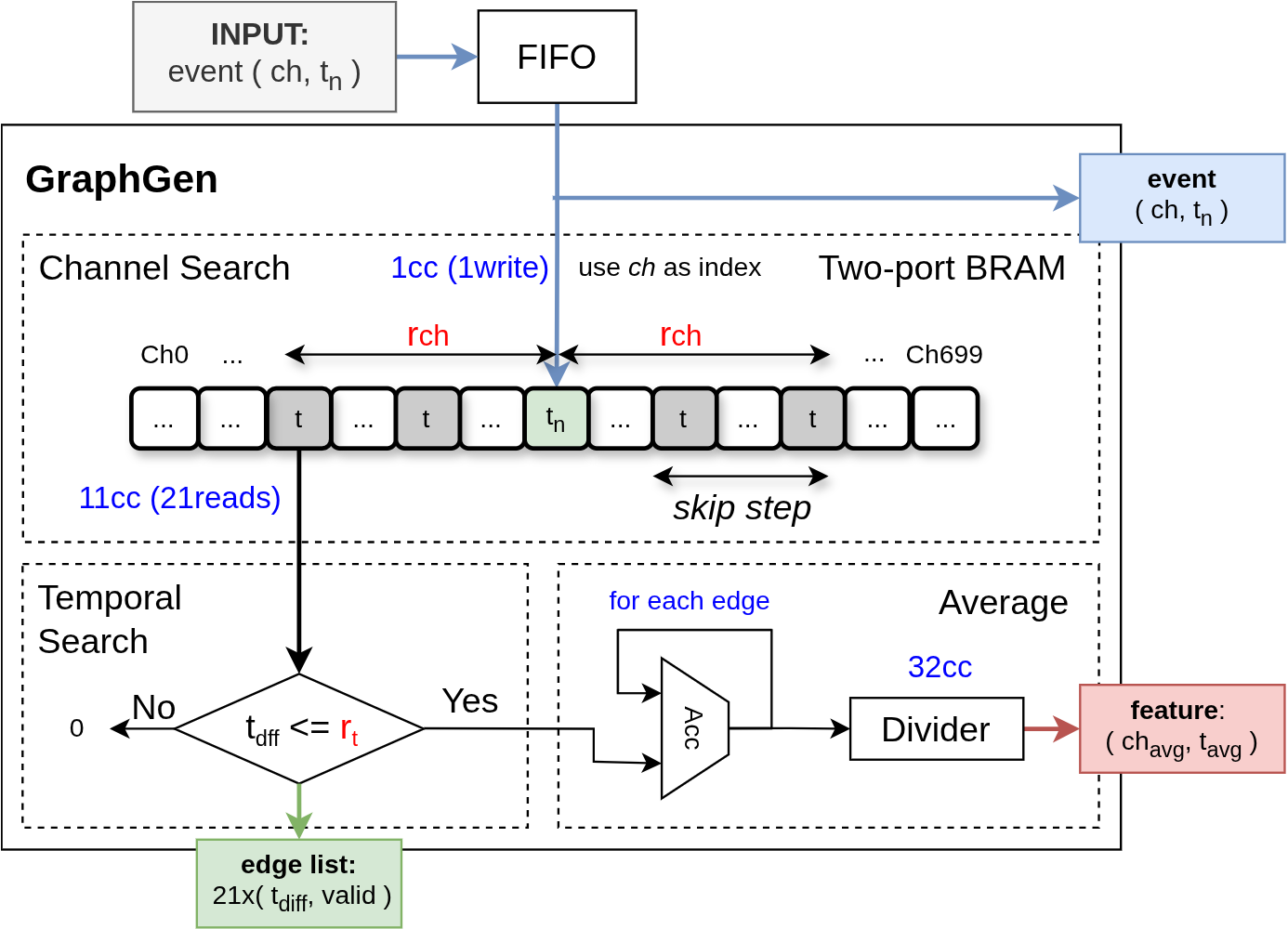}
    \caption{Diagram of the graph generation hardware module, which takes events from FIFO as input, and outputs the vertex features along with the list of its edges. The blue colour indicates the time (measured in clock cycles) required for the individual steps.}
    \Description{The figure shows the hardware graph generation module. Events are read from a~FIFO buffer, searched across channels and temporal windows, and stored in BRAM. Valid edges are identified through temporal search, while features are computed using an accumulator and divider. The output consists of vertex features and the corresponding edge list, with blue labels indicating clock cycle costs of each step.}
    \label{fig:graph-gen}
\end{figure}

Figure~\ref{fig:graph-gen} illustrates the main components and data flow of the graph generation module. Asynchronous events, defined by their time \( t \) and channel index \( ch \), enter a~first-in first-out (FIFO) buffer, ensuring a~stable data stream. Each event then moves to a~1D context memory indexed by the channel.

The module performs a~neighbour search using the \textit{skip step connection} method. 
Instead of scanning all possible neighbours, the system reads events at fixed intervals along the channel dimension from the BRAM. 
Events that meet the \textit{temporal search} criteria (based on \( r_t \) radius) are added to an edge list, and their corresponding time and channel indices are fed into the accumulator.
After iterating over all channels, this accumulator computes the average position of the neighbours using a~simple divider.
The resulting averaged features, together with the newly stored event and the edge list, are then passed to the first stage of the graph convolution pipeline.

The total number of clock cycles \( N_{\text{cycles}} \) required for graph generation depends on the channel radius (\( r_{ch} \)), the \textit{skip step} (\( s \)), and the additional cycles for feature computation (\( N_{\text{div}} \)). Assuming dual-port memory (two reads per clock cycle), this can be expressed as:
\begin{equation}
N_{\text{cycles}} = \frac{1 + 2 \cdot \frac{r_{ch}}{s}}{2} + N_{\text{div}}
\end{equation}

Here, \(1\) corresponds to the central channel read, \(2 \cdot \frac{r_{ch}}{s}\) represents the reads from the upper and lower channels, and \( N_{\text{div}} \) accounts for division. 
\changed{In our implementation (designed based on ablation studies described in Section \ref{sec:ablation}) we selected the radius $r_{ch}$ of 100 with \textit{skip step} $s$ of 10.}

\subsubsection{\textbf{Graph convolution}}
\label{sec:convolution}

In the baseline approach, the PointNetConv \cite{POINTNET} convolutional layer was used, which extends the classical \textit{message passing mechanism} commonly used in graph convolutions. 
As highlighted in \cite{jeziorek2023memory,jeziorek2024embedded}, these layers are lightweight and well-suited for hardware acceleration, making them a~natural choice for adoption in our work. However, we introduce two key modifications.


First, we integrate a~\textit{batch normalisation (BN)} layer into \(\phi\) to ensure stable training. Notably, during quantisation these layers are folded \cite{jacob2018quantization}, which prevents any increase in the model parameters for hardware deployment. The second and more important change involves an additional normalisation step for the positional differences used on the event-graph edges. The input data is initially normalised to \((0, 1)\) range, while neighbours are determined within the radius \(r_{ch}\) and \(r_t\). Consequently, channel index differences are within \((-r_{ch}, r_{ch})\), and time index differences are within \((-r_t, 0)\) due to the use of a~time-based directed graph generator. These values occupy only a~small fraction of \((0, 1)\) range, which adversely affects training and reduces precision during quantisation. To address this, we apply \textit{positional normalisation (PN)} after computing position differences, rescaling the values back to \((0, 1)\) range, by multiplying time differences by \(-\frac{1}{r_t}\), and channel differences by adding \(r_{ch}\) and multiplying by \(\frac{2}{r_{ch}}\).

The modified model is expressed as:

\begin{equation}
\label{eq:pointnetconv1}
\hat{\mathcal{X}_i} = \max_{j \in N(i)} (BN\ \phi([\mathcal{X}_j || PN(\mathcal{P}_j - \mathcal{P}_i)])).
\end{equation}

\begin{figure}[t]
    \centering
\includegraphics[width=0.8\linewidth]{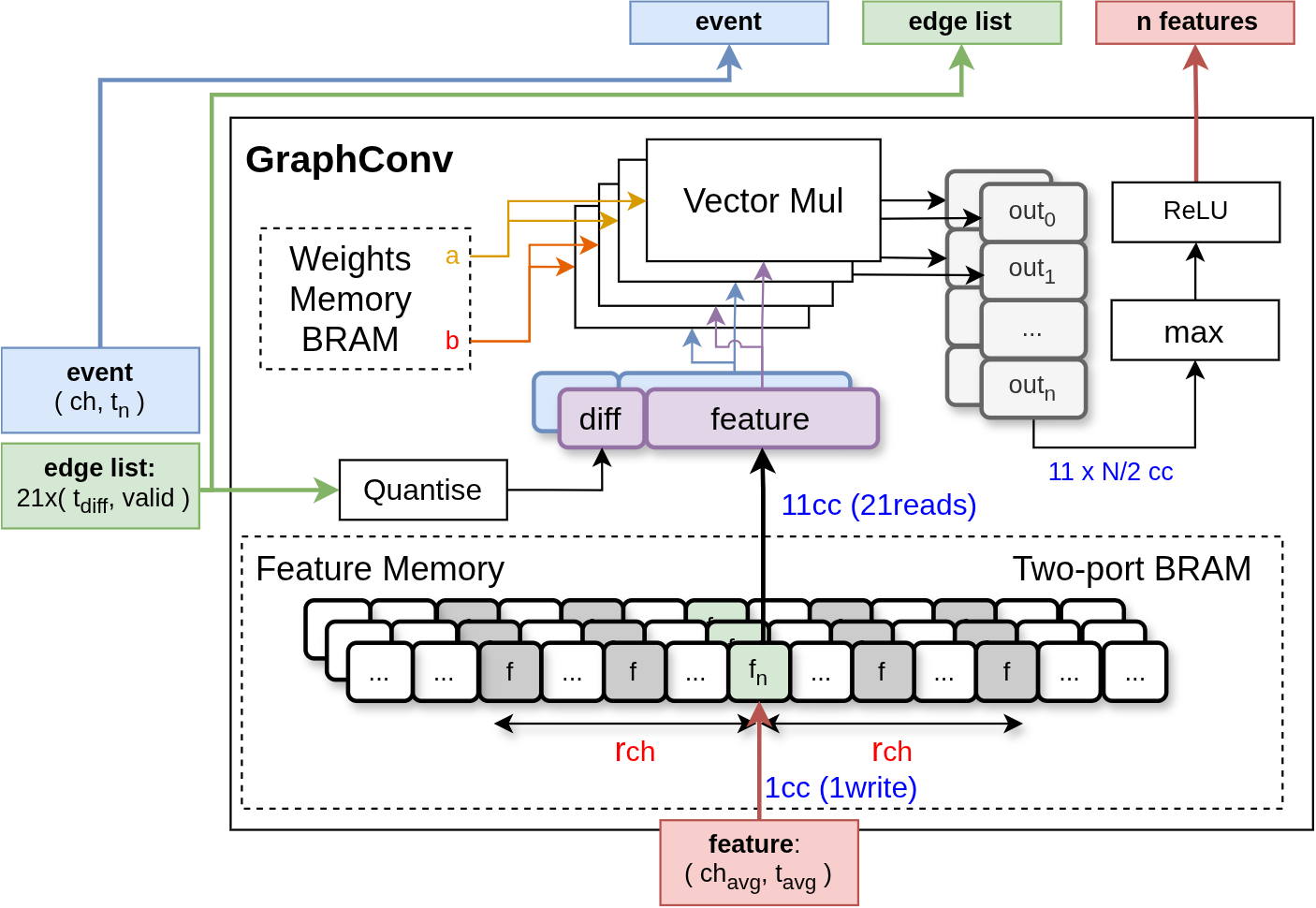}
    \caption{Diagram of the graph convolution hardware module. The arrows indicate the dataflow in the module, while the blue text indicates the time (measured in clock cycles) required for the individual steps.}
    \Description{The figure illustrates the hardware architecture of the graph convolution (GraphConv) module. Events and their edge lists are processed using stored features in two-port BRAM. Features are quantised and combined with edge differences, then multiplied with weights from memory. The results undergo ReLU activation and max pooling to generate output features. The diagram also shows clock cycle costs for read and write operations, highlighting the efficiency of the hardware design.}
    \label{fig:graph-conv}
\end{figure}

For hardware implementation each graph convolution assumes fully asynchronous event-by-event processing and can be executed in parallel.
Each incoming event with its edge list and input features is processed independently. 
The module must also access the features of each vertex connected with an edge and the difference between the neighbour and the processed event.
For this purpose, we implemented one 2-port BRAM memory that stores the features of the last processed event for each channel, and another one for storing the weights (see Figure \ref{fig:graph-conv}).

To determine the final output feature, the linear layer must be applied for the vertex itself (so-called self-loop) and for each of its neighbours (a~maximum of \texttt{MAX\_EDGE} = 21 times with the \(r_{ch}\) = 100 and \textit{skip step} \(s\) = 10). 
The key part of the graph convolution is the vector multiplication module -- in order to reduce resource utilisation, the module was implemented in a~way that some of the calculations were performed sequentially (inspired by \cite{jeziorek2024embedded}). 

Taking advantage of two-port memories, we process two feature vectors and calculate two elements of the output vector at the same time (cf. Figure \ref{fig:graph-conv}). The total number of clock cycles required to perform the convolution for a~single event can be determined with the following formula:

\begin{equation}
\label{eq:pointnetconvHW}
N_{\text{cycles}} = \frac{MAX\_EDGE+1}{2}\cdot \frac{OUT\_DIM}{2} 
\end{equation}

The graph convolution module can be considered a~bottleneck of the proposed method. The throughput of the system is dependent on the maximum number of clock cycles required per single event. For our baseline model designed for classification and implemented for a~200 MHz clock, the theoretical throughput was thus calculated to be 555 kEPS (thousand events per second). At the same time, the average value in the SHD dataset is around 20 kEPS. 
However, it should be noted that the number of parallel vector multiplication modules could be increased to improve latency and throughput (using additional resources), or decreased with an opposite effect.
\changed{This flexibility improves system scalability and applicability for different tasks and requirements.}

An essential part of the system is the applied quantisation. In our work, \changed{all multiplications are performed exclusively on integer values} with the bit precision that can be selected per layer. 
Both feature map elements and weights are stored in memory as unsigned integers, and rescaled before (quantisation) and after (re-quantisation) multiplications. 
For scaling, we use DSP multiplication and bit-shifting or look-up tables (depending on the number of possible quantised values) adhering to the methodology proposed in \cite{jacob2018quantization}.

\subsection{Classification}


\changed{The first task addressed in this work is time-series classification. In this setting, an event-based data stream corresponding to the recording of a~single word is provided to the system. The prediction is generated only after the entire sample has been fully processed. To achieve this, the event features are first aggregated using graph average pooling, and the resulting representation is then passed to a~classifier for the final score.}

\subsubsection{\textbf{Graph average pool}}
\label{sec:averagepool}


\changed{To obtain a~compact representation of the event features produced by the graph convolution layers, a~global graph pooling operation is applied. This step aggregates information from all events within a~sample, reducing the representation to a~single fixed-length vector that can be processed by fully connected layers. In the original approach \cite{lars}, global average pooling was employed. By averaging the accumulated event features, the model captures a~global summary that is particularly suitable for subsequent classification tasks. In our experiments (Section \ref{sec:graph_pooling}), we tested simpler alternatives such as global maximum pooling and global sum pooling; however, these methods led to a~notable performance drop (by 7 pps for max pooling and 5 pps for sum pooling). Based on these findings, we retain the original average pooling strategy.}

We implemented a~global average pooling module which receives the features from the last stage of the graph convolution (Figure \ref{fig:overview}) to the input. 
The mean value is calculated by accumulating the sum of each vector feature in a~given register.
To determine the number of events that have been accumulated, a~simple counter is used.
When the output vector corresponding to the last event in a~sample is calculated, the accumulator register values are divided by the current value of the counter. These values are then output by the module and stored in a~BRAM memory. 

\subsubsection{\textbf{Classification head}}
\label{sec:classhead}

\changed{The pooled feature vector obtained from the global graph pooling layer is subsequently passed to a~classification head. This module consists of a~multi-layer perceptron (MLP), designed to map the compact representation into the target class space. The MLP applies two fully connected layers with ReLU activation functions. A~softmax function is then used to normalise the output scores into class probabilities.

For FPGA-based hardware, we investigated two approaches for implementing the classification head. The first approach leverages the heterogeneous nature of the target platform by executing the multilayer perceptron within the processing system (PS) rather than the in programmable logic (PL). Since the MLP is applied only after the entire input sequence has been processed by the graph neural network feature extractor, hardware acceleration is not critical at this stage. In this configuration, the MLP operates with full precision using floating-point arithmetic, without quantisation.

However, communication between the PS and PL introduces additional latency, and execution on the CPU increases energy consumption. To address these limitations, we also implemented the classification head within the reconfigurable PL. In this design, the MLP weights and activations are quantised to 8 bits, and computation is carried out using two vector-multiplication modules. This architecture was specifically chosen to efficiently exploit the double-port BRAM memory available for weight storage.
A~detailed evaluation and comparison of these two approaches is presented in Section \ref{sec:hw_scalablity}.}

\subsection{Keyword spotting}

\changed{While the simple classification task assumes that the entire word is contained within a~single input sample, keyword spotting requires operating on a~continuous event stream, where the offset of a~word is not known in advance. In this scenario, the system must both detect the temporal boundaries of a~spoken keyword and classify its content.

To address this challenge, the event stream is divided into smaller temporal windows. For each window, event features are extracted and pooled. These window-level features are then processed by a~recurrent module, enabling the system to capture long-term temporal dependencies and to make predictions in an online manner.}

\subsubsection{\textbf{Graph max pool}}
\label{sec:maxpool}

\changed{In contrast to the classification task, in keyword spotting the aggregation of events is performed over a~significantly smaller temporal window $\Delta t$ (1000 ms for classification vs 10 ms for KWS). As a~result, the use of an element-wise operator has a~less detrimental effect on performance. As demonstrated in the ablation studies (Section \ref{sec:graph_pooling}), the difference between applying average/add pooling and max pooling under these conditions is negligible, with a~slight advantage for the max operation. Additionally, for hardware implementation, max pooling is preferable, since it eliminates the need for accumulation and division, and reduces the operation to a~simple element-wise comparison and replacement. Therefore, in the keyword spotting task we adopt graph max pooling.}

\subsubsection{\textbf{KWS head}}
\label{sec:kwshead}

\changed{The KWS head operates on a~stream of pooled feature vectors computed over fixed-length windows of size $\Delta t$.
Let $\{\mathbf{f}_t\}_{t=1}^T$ denote the sequence of window-level vectors (one per $\Delta t$ window). 
Each $\mathbf{f}_t$ is first transformed by a~lightweight two-layer STEM with nonlinearities (ReLU), which improves local representation for temporal modelling.
The STEM is intentionally shallow to minimise latency and resource usage.

For temporal modelling we employ a~gated recurrent unit (GRU) \cite{gru}, chosen as a~compromise between model quality and resource utilisation. 
Vanilla RNNs are prone to vanishing gradients and insufficient temporal context modelling, while long short-term memory (LSTM) \cite{lstm} networks, although often more accurate, incur higher parameter counts and latency due to an additional gate and explicit cell state updates. 

Let $\mathbf{x}_t$ be the STEM-transformed input at time $t$, and $\mathbf{h}_{t-1}$ the previous hidden state. The GRU updates are:
\begin{align}
\mathbf{z}_t &= \sigma\!\left(W_z \mathbf{x}_t + U_z \mathbf{h}_{t-1} + \mathbf{b}_z\right), \\
\mathbf{r}_t &= \sigma\!\left(W_r \mathbf{x}_t + U_r \mathbf{h}_{t-1} + \mathbf{b}_r\right), \\
\tilde{\mathbf{h}}_t &= \tanh\!\left(W_h \mathbf{x}_t + U_h \big(\mathbf{r}_t \odot \mathbf{h}_{t-1}\big) + \mathbf{b}_h\right), \\
\mathbf{h}_t &= (1-\mathbf{z}_t)\odot \mathbf{h}_{t-1} + \mathbf{z}_t \odot \tilde{\mathbf{h}}_t,
\end{align}
\noindent where $\sigma(\cdot)$ denotes the logistic sigmoid function, $\odot$ represents element-wise multiplication, and ${W_\cdot, U_\cdot, \mathbf{b}_\cdot}$ are the learnable weight matrices and bias vectors associated with the reset gate ($r$), update gate ($z$), and candidate hidden state ($h$), respectively.
The hidden state $\mathbf{h}_t$ integrates features over time and is updated once per $\Delta t$ window, enabling strictly online operation.

At each time step $t$, two outputs are produced from the current hidden state $\mathbf{h}_t$:
(i) class probabilities (\textit{CLS}), obtained through a~linear transformation followed by a~softmax activation, and
(ii) a~scalar confidence score (\textit{CONF}), obtained through a~linear transformation followed by a~sigmoid activation.
Consequently, every $\Delta t$ window yields both a~class prediction and an associated confidence estimate.

Given the significantly lower pooling times compared to the classification task (10 ms vs 1000 ms), efficient computation of the final prediction becomes highly desirable. Based on this observation, we implemented the KWS head in the programmable logic alongside feature extraction and pooling, in order to minimise latency and energy consumption.

\begin{figure}[t]
    \centering
\includegraphics[width=0.8\linewidth]{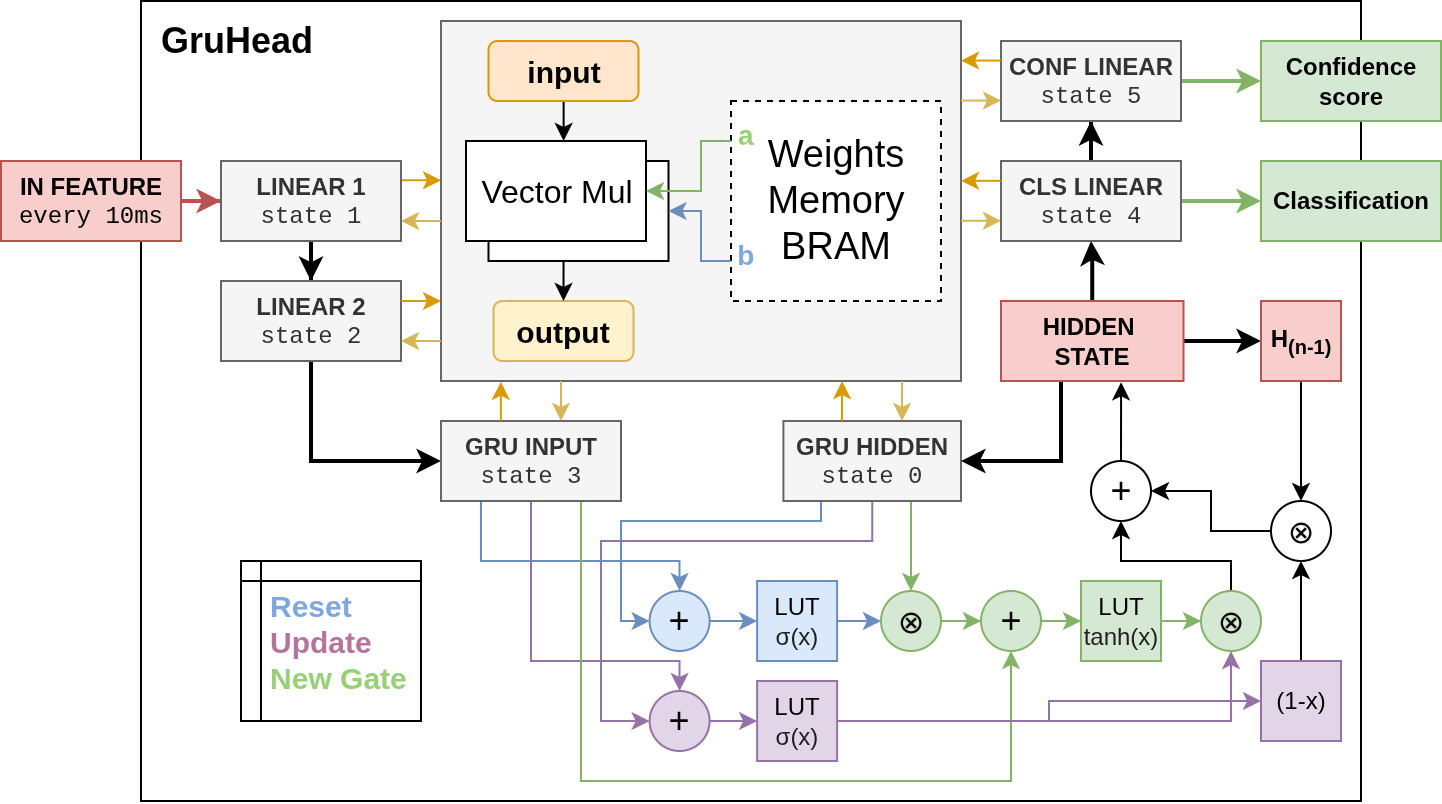}
    \caption{Diagram of the head of the hardware KWS module. The key part of this module is vector multiplication shared between consecutive MLPs -- its utilisation is controlled with the state machine.}
    \Description{The figure shows the head of the hardware keyword spotting module (GruHead). Input features, arriving every 10 ms, are processed by linear layers and vector multiplications with weights stored in BRAM. A~GRU cell updates hidden states using lookup-table-based activation functions (sigmoid and tanh), controlled by a~state machine. The outputs are passed through linear layers for classification and confidence score estimation, enabling efficient keyword spotting.}
    \label{fig:gru-head}
\end{figure}

The head of the hardware KWS module is organised as a~state machine that sequentially controls the computations required for prediction (illustrated in Figure~\ref{fig:gru-head}). Unlike the rest of the system, which is fully pipelined and optimised for parallel execution, this component employs a~control-flow–oriented design. The core of the module consists of two vector multiplication units. They are used to efficiently exploit the dual-port memory available for weight storage; however, the level of parallelism can be adjusted -- more units can be instantiated to reduce latency, while fewer units can be used to save energy and FPGA resources. Each matrix multiplication required for prediction is performed sequentially (processing two rows in parallel), with the state machine orchestrating input reloading and output accumulation.

Using this scheme, the first and second MLPs of the STEM are computed in states 1 and 2, respectively. The hidden state $\mathbf{h}_t$ is then updated according to the GRU recurrence equations, based on the previous hidden state $\mathbf{h}_{t-1}$ and the STEM-transformed input $\mathbf{x}_t$. GRU-related multiplications (state 3) reuse the same resources as the STEM computations. Nonlinear functions, including the $sigmoid$ and $tanh$, are implemented via look-up tables, while the Hadamard product is mapped to the FPGA’s logic fabric. The updated hidden state is then passed through the classification and confidence score layers, again reusing the same multiplication resources (states 4 and 5). At this stage, the final prediction is generated and forwarded to the module output.

Finally, the hidden state is also supplied to the GRU MLP to compute the term $U\mathbf{h}_{t-1} + \mathbf{b}$ required for the next iteration (state 0). This step is also triggered once at the beginning of processing for initial $\mathbf{h}_{t-1}$ (hence the state ``0'').
After each multiplication in the KWS head module, we apply re-quantisation following the methodology described in \cite{jacob2018quantization}, implemented using the FPGA’s DSP resources. 

The proposed head implementation was designed to minimise resource utilisation. Since this operation is executed only once every 10 ms, the additional latency introduced by the sequential computations does not impose a~significant delay on the overall system.}

\subsubsection{\textbf{Data preparation}}
\label{sec:kws_preparation}

\begin{figure}[t]
    \centering
    \begin{subfigure}{0.28\textwidth}
        \centering
        \includegraphics[width=\linewidth]{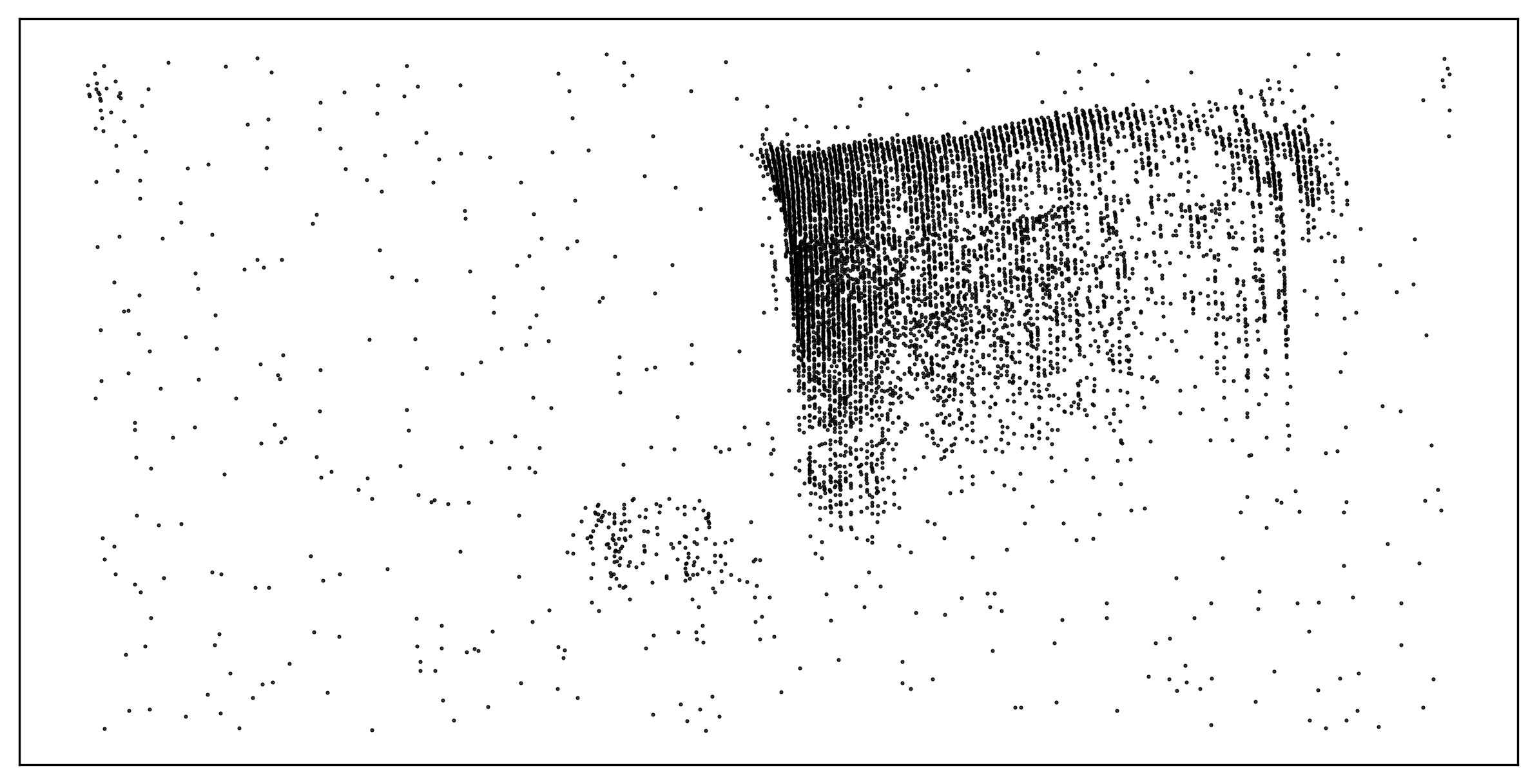}
    \end{subfigure}
    \hspace{0.7cm}
    \begin{subfigure}{0.28\textwidth}
        \centering
        \includegraphics[width=\linewidth]{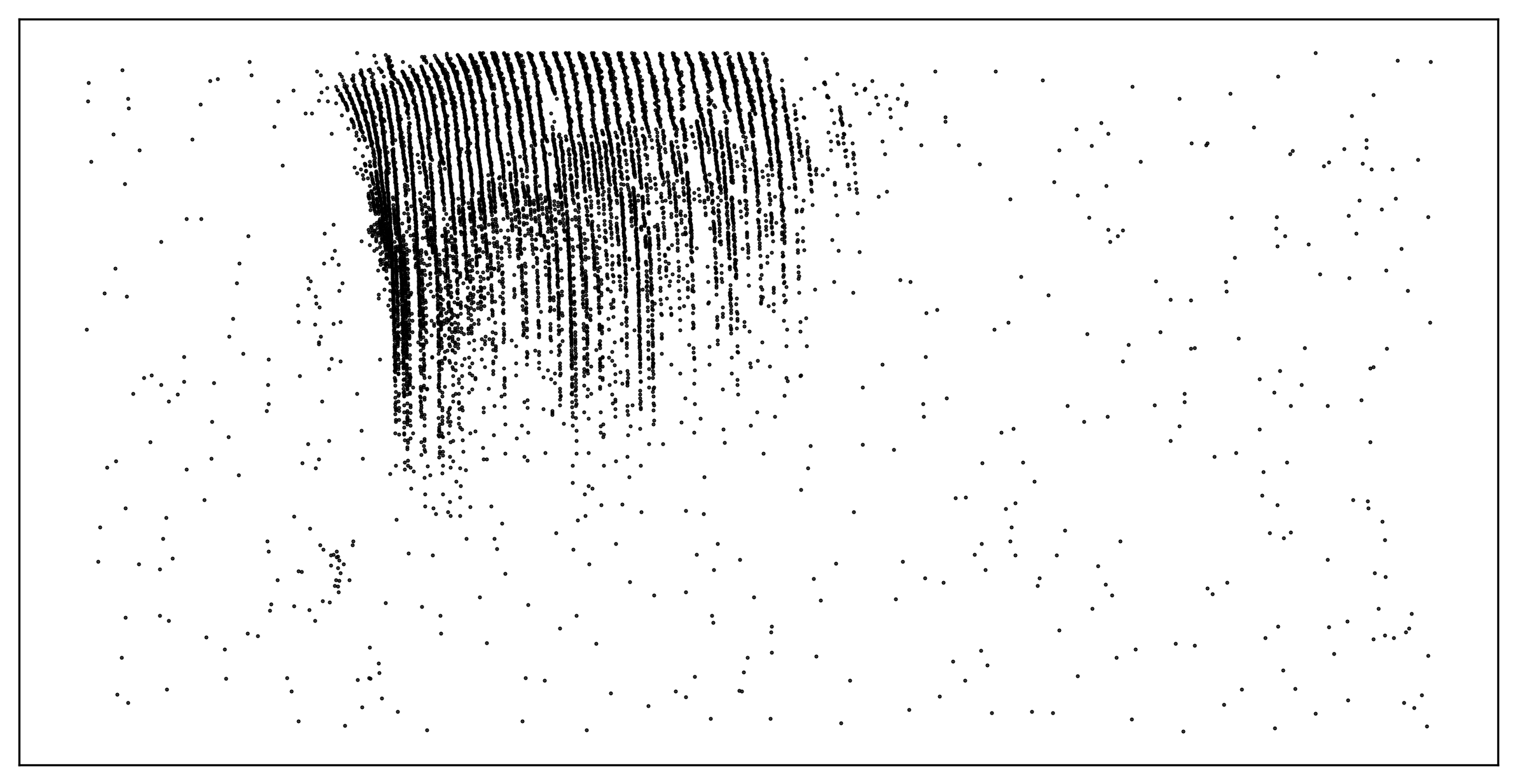}
    \end{subfigure}
    \hspace{0.7cm}
    \begin{subfigure}{0.28\textwidth}
        \centering
        \includegraphics[width=\linewidth]{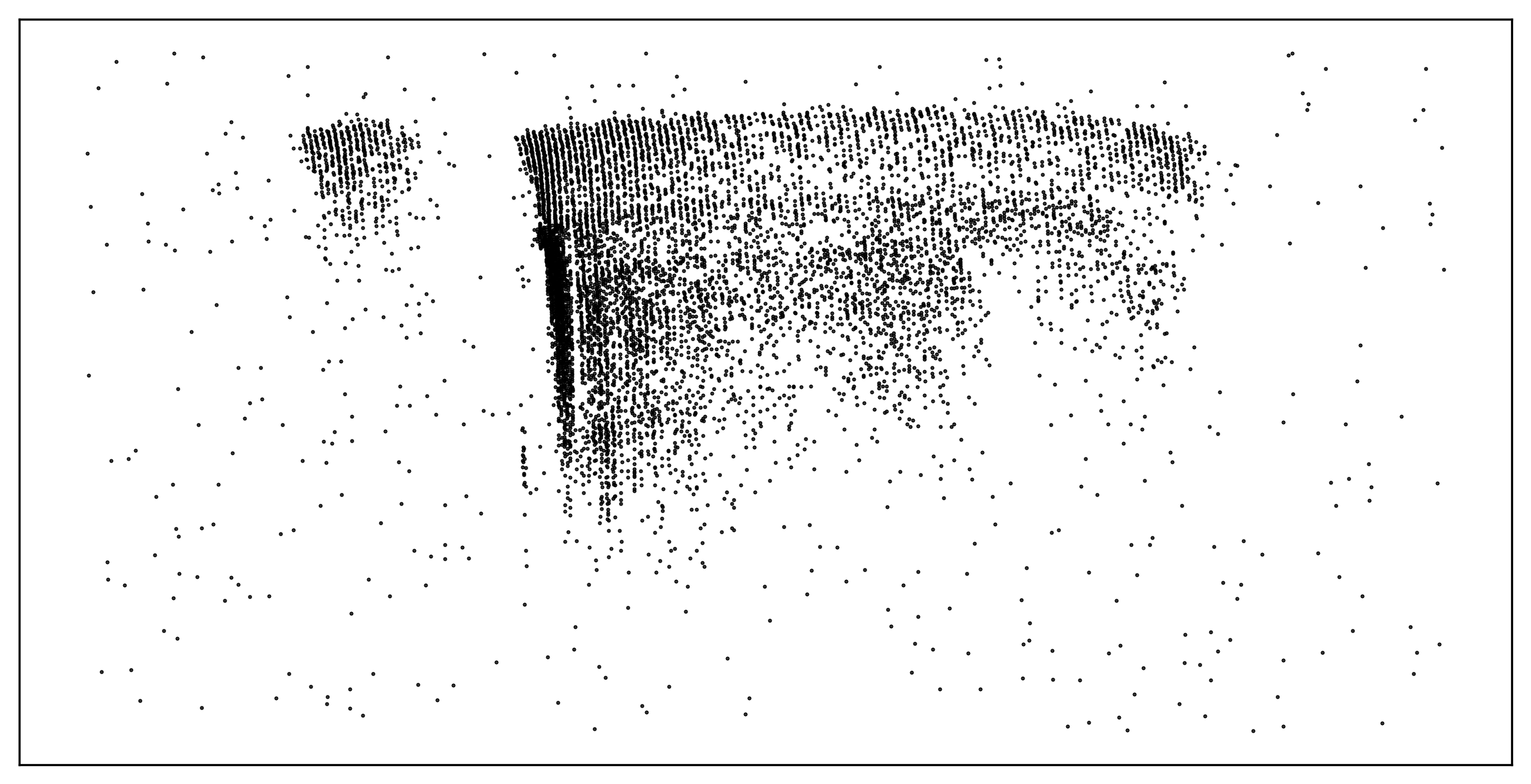}
    \end{subfigure}

    \vspace{0.3cm}

    \begin{subfigure}{0.33\textwidth}
        \centering
        \includegraphics[width=\linewidth]{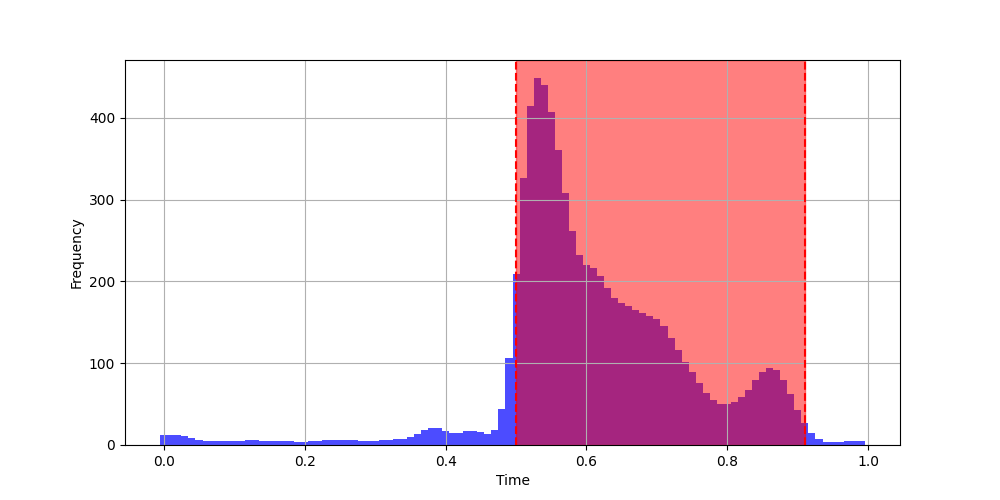}
    \end{subfigure}
    \begin{subfigure}{0.33\textwidth}
        \centering
        \includegraphics[width=\linewidth]{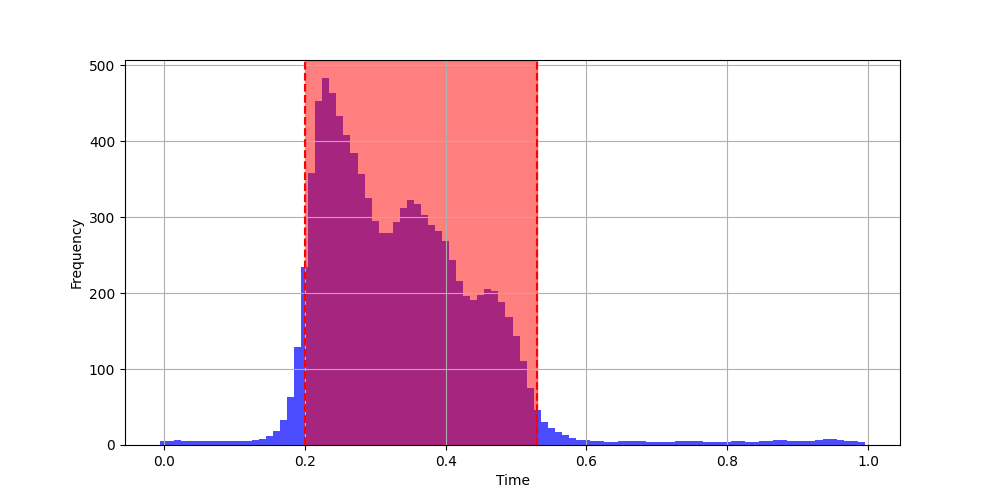}
    \end{subfigure}
    \begin{subfigure}{0.33\textwidth}
        \centering
        \includegraphics[width=\linewidth]{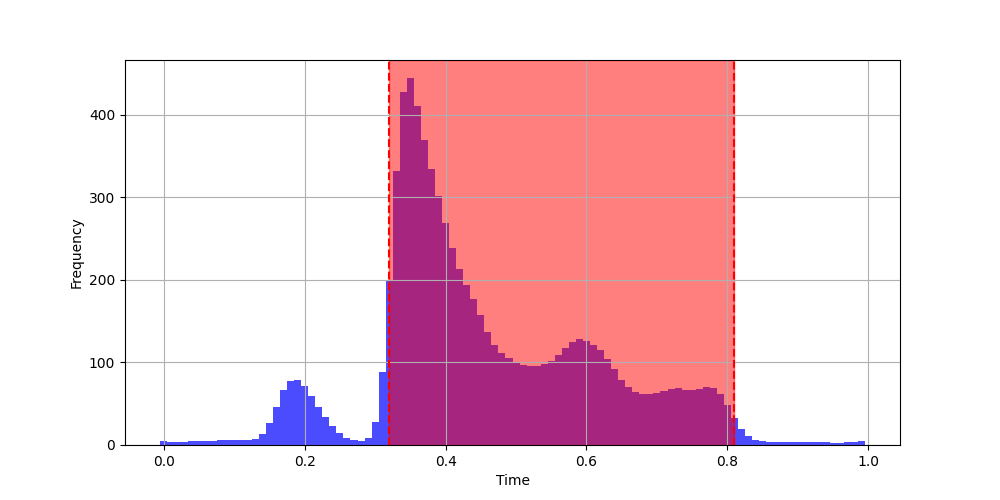}
    \end{subfigure}

    \caption{Examples of spiking event stream with automatically extracted keyword boundaries. Top: raw spiking events; bottom: smoothed histogram with detected active range.}
    \Description{The figure presents examples of spiking event streams with extracted keyword boundaries. The top row shows raw spiking events over time, while the bottom row shows corresponding smoothed histograms. The highlighted red regions indicate the automatically detected active keyword ranges.}
    \label{fig:events_hist}
\end{figure}

\changed{The original SHD and SSC datasets were designed for recognition tasks and therefore do not include time-aligned onset/offset annotations for individual keywords. To obtain labels suitable for keyword spotting, we derive segment boundaries from the continuous event stream by thresholding a~\textit{smoothed activity histogram} constructed on a~\textit{fixed temporal grid} with bin width $\Delta t$ over a sample time window $T$.

Let $T>0$ denote the per-sample duration, and let $\Delta t>0$ be the histogram bin width (in milliseconds). We form $N=\left\lceil T/\Delta t \right\rceil$ bins with boundaries:

\begin{equation}
\texttt{bin\_edges}=\big[0,\ \Delta t,\ 2\Delta t,\ \ldots,\ N\Delta t\big]\cap[0,T],
\end{equation}

\noindent and define $\texttt{hist}\in\mathbb{R}^{N}$ as event counts per bin. Our objective is to estimate the active interval $[t_{\text{start}},\,t_{\text{end}}]\subseteq[0,T]$ of the spoken keyword, which we then use as ground truth for KWS training and evaluation.

We first obtain a~smoothed signal by convolving the raw histogram with a~1-D Gaussian kernel of length $k$ in bins, which suppresses noise and micro-pauses:

\begin{equation}
\texttt{hist}^{(\text{smooth})} = \texttt{hist} * g_\sigma,\qquad 
g_\sigma[m] \propto \exp\!\big(-\tfrac{m^2}{2\sigma^2}\big),\ \ m\in\{-\tfrac{k-1}{2},\ldots,\tfrac{k-1}{2}\}.
\end{equation}

On this smoothed trace, we compute \textit{adaptive hysteresis thresholds}. The high threshold for activation is:

\begin{equation}
T_{\text{high}}=\mu+\alpha\sigma,\quad
\end{equation}

\noindent where $\mu$ and $\sigma$ are the mean and standard deviation of $\texttt{hist}^{(\text{smooth})}$. The low threshold is a~fraction of the high threshold, $T_{\text{low}}=\beta\,T_{\text{high}}$. This hysteresis gap stabilises detection across varying recording levels and mitigates rapid switching near the decision boundary.

Activation onset is defined as the first upward crossing of $T_{\text{high}}$. 
Once activated, the state remains active until $\texttt{cooldown\_steps}$ consecutive bins fall below $T_{\text{low}}$, 
implementing a~cooldown of duration: 
\begin{equation}
\tau_{\text{cool}} = \texttt{cooldown\_steps} \cdot \Delta t,
\end{equation}
which prevents a~single keyword from being fragmented into multiple segments due to brief dips in activity. 
The deactivation time $t_{\text{end}}$ is assigned to the corresponding bin edge. 
If the sequence reaches the end of the recording at $T$ while still active, we set $t_{\text{end}} = T$. 
Under the single-keyword assumption, only the first such active interval is retained. 
Samples without any upward crossing of $T_{\text{high}}$ are labelled as ``no-keyword'' (or excluded from training). 
Finally, extremely short detections are suppressed by enforcing a~minimum duration:
\begin{equation}
(t_{\text{end}} - t_{\text{start}}) \geq \delta_{\min},
\end{equation}
equivalently $\geq \lceil \delta_{\min}/\Delta t \rceil$ bins.

The resulting automatically derived onset/offset labels -- visualised in Figure~\ref{fig:events_hist} -- are used as ground truth for training and evaluating our KWS models. 
The best-performing hyperparameter settings, empirically validated for both SHD and SSC datasets, are summarised in Table~\ref{tab:kws_hparams}.}

\begin{table}[t]
\centering
\caption{Hyperparameters for $\Delta t$-binned KWS label extraction.}
\label{tab:kws_hparams}
\resizebox{\textwidth}{!}{%
\begin{tabular}{@{}llll@{}}
\toprule
\textbf{Quantity}                       & \textbf{Symbol/name}           & \textbf{Default/example} \\ \midrule
Sample window                  & $T$                     & 1 s (based on the sample)        \\
Histogram bin width            & $\Delta t$ 		 & 10 ms                              \\
Gaussian kernel size (in bins) & $k$                       & 7                                  \\
High threshold                 & $T_{\text{high}}=\mu+\alpha\sigma$                   & $\alpha=0.5$                             \\
Low threshold                  & $T_{\text{low}}=\beta T_{\text{high}}$                    & $\beta=0.2$                              \\
Cooldown                       & \texttt{cooldown\_steps}, $\;\tau_{\text{cool}}=\texttt{cooldown\_steps}\cdot\Delta t$                      & 5 bins                          \\
Min. duration                  & $\delta_{\min}$                   & $40$ ms             \\ \bottomrule
\end{tabular}%
}
\end{table}



\section{Evaluation}
\label{sec:evaluation}

\subsection{Setup}
\label{sec:setup}

All investigated models share a~common feature-extraction backbone composed of four PointNetConv layers followed by a~graph pooling layer (Section~\ref{sec:proposed_system}).
For the classification task, the backbone is extended with a~two-layer multilayer perceptron classifier (Section \ref{sec:classhead}).
For the keyword spotting (KWS) task, the backbone is augmented with a~two-MLP STEM preceding a~gated recurrent unit (GRU) layer (Section \ref{sec:kwshead}). Two task-specific output heads are then attached: (i) a~confidence head, which estimates the most likely onset timestep of the keyword within the input sequence, and (ii) a~classification head, which assigns the detected segment to one of the $C$ predefined classes. \changed{Taking inspiration from prior works that use the Google Speech Commands dataset for command classification, we divided the Spiking Speech Commands dataset into two subsets: SSC-35 and SSC-11. In SSC-35, all classes are considered relevant, while in SSC-11, only 10 target words (e.g. stop, go, off, right) are retained, with the remaining words are grouped into an additional class labelled ``unknown''.}

All models were implemented in PyTorch \cite{paszke2017automatic} with the PyTorch Lightning framework \cite{Falcon_PyTorch_Lightning_2019}.
For the classification task, the training was performed using a~standard cross-entropy loss applied to the classifier outputs.
For the KWS task, a~multitask learning strategy was employed. The confidence head was trained using a~weighted binary cross-entropy loss over all timesteps, where the positive class was weighted by the ratio of negative to positive samples ($99:1$) to mitigate severe class imbalance. The classification head was trained exclusively at the timestep corresponding to the maximum confidence score, using a~cross-entropy loss scaled by a~factor of 5.0. The overall training objective was defined as a~weighted sum of these two components.

All models were trained for 100 epochs in full-precision (FP32), followed by an additional 20 epochs of quantisation-aware training (QAT) \cite{jacob2018quantization}. A~batch size of 16 was used for classification and 4 for KWS. The training was conducted on a~single NVIDIA A100 GPU. The Adam optimiser \cite{kingma2014adam} was employed with a~learning rate of $2\cdot10^{-4}$, weight decay of $1\cdot10^{-4}$, and a~ReduceLROnPlateau scheduler with a~reduction factor of 0.5 and patience of 10 epochs. Model checkpoints were selected based on the lowest training loss for the SHD and evaluation loss for the SSC and subsequently evaluated on the test set. 

For evaluation, we employed task-specific metrics. In classification, the primary metric was the top-1 accuracy. In KWS, two complementary metrics were used:

\begin{enumerate}
    \item \textbf{Top-1 accuracy at maximum confidence} ($Acc_{\text{K}}$), defined as the classification accuracy at the timestep with the highest predicted confidence.
    \item \textbf{Tolerance-aware top-1 accuracy} ($Acc_{\text{K,$\Delta$}}$), which extends $Acc_{\text{K}}$ by considering predictions correct if the maximum-confidence timestep falls within a~tolerance window of $\pm 1$ temporal bin from the ground truth onset.
\end{enumerate}



The hardware architecture of the network was implemented in the SystemVerilog language using Vivado 2022.2, while the processing system was programmed using Vitis 2022.2 in the C++ language.
After verifying the compatibility with PyTorch software model via Vivado simulation, it was implemented on a~Zynq UltraScale+ ZCU104 FPGA platform with XCZU7EV chip from AMD Xilinx.
\changed{
As we do not use a~physical AC sensor for hardware testing, we employ the FPGA’s processing system to read events from an SD card and feed them to the programmable logic according to their timestamps, thereby simulating the temporal sparsity of the data.
Based on experiments conducted with reconfigurable logic, we selected a~clock frequency of 200 MHz. This choice ensures low latency while avoiding timing issues during system implementation.}

\subsection{Ablation studies}
\label{sec:ablation}

In this section, we present the results of experiments conducted to evaluate the influence of hyperparameters and selected operations on the proposed architecture. We begin by analysing the impact of the graph generator parameters. Next, we compare the performance and parameter efficiency of different model variants to assess the scalability of the approach. Finally, we evaluate the effect of max, mean, and average operations in graph pooling for both classification and keyword spotting tasks.
Unless stated otherwise, the base model was configured with all output channels set to 64, a~graph generator with an \(r_{ch}\) of 100, a~\textit{skip step} of 10, an \(r_t\) of 20 ms \changed{and using the SHD dataset}.

\subsubsection{\textbf{Graph generation configurations}}

In Table \ref{tab:graph_gen} we present the results of our accuracy analysis for the base model as a~function of the \(r_{ch}\) and \textit{skip step} parameters. The experiment was carried out in two stages. In the first one, we varied the \(r_{ch}\) parameter from 30 to 300 while keeping the \textit{skip step} fixed at 10. In the second stage, using the best \(r_{ch}\) value identified in the first stage, we varied the \textit{skip step} parameter from 1 to 20. A~``–''~score indicates configurations excluded from the analysis due to excessively large tensor sizes that could not fit in the GPU memory.

\begin{table}[t]
\caption{The impact of search radius \(r_{ch}\) and \textit{skip step} parameters on top-1 accuracy. The best results, highlighted in \textbf{bold}, were obtained for moderate parameters.}
\resizebox{0.8\textwidth}{!}{%
\begin{tabular}{@{}cccc|cccc@{}}
\toprule
~~\(r_{ch}\)~~ &
  ~~\textit{skip step}~~ &
  \begin{tabular}[c]{@{}c@{}}~~top-1 acc.~~\\ float\end{tabular} &
  \begin{tabular}[c]{@{}c@{}}~~top-1 acc.~~\\ quantised\end{tabular} &
  ~~\(r_{ch}\)~~ &
  ~~\textit{skip step}~~ &
  \begin{tabular}[c]{@{}c@{}}~~top-1 acc.~~\\ float\end{tabular} &
  \begin{tabular}[c]{@{}c@{}}~~top-1 acc.~~\\ quantised\end{tabular} \\ \midrule
100 & 1  & 90.95 \% & -        & 30  & 10 & 85.43 \% & 84.23 \% \\
100 & 5  & 91.37 \% & 90.10 \% & 50  & 10 & 88.25 \% & 87.60 \% \\
\textbf{100} & \textbf{10} & \textbf{92.74 \%} & \textbf{92.30 \%} & 200 & 10 & 91.23 \% & 90.05 \% \\
100 & 15 & 90.63 \% & 89.22 \% & 250 & 10 & 90.47 \% & 89.72 \% \\
100 & 20 & 90.05 \% & 88.91 \% & 300 & 10 & 89.66 \% & 88.62 \% \\ \bottomrule
\end{tabular}%
\label{tab:graph_gen}
}
\end{table}

We observed that too few edges -- caused by either a~high \textit{skip step} or a~low \(r_{ch}\) -- negatively affect the model accuracy. Conversely, an excessively large \(r_{ch}\) or too small \textit{skip step} can also degrade classification performance. This is due to the sensitivity of PointNetConv layers to outliers, which become more noticeable with a~large number of edges -- a~consequence of using a~max-type aggregation. The best results were achieved with moderate parameter values: an \(r_{ch}\) of 100 and a~\textit{skip step} of 10, yielding 92.74\% accuracy for the floating-point model. Although it is possible that simultaneously increasing both the \(r_{ch}\) and the \textit{skip step} might lead to even better performance, the chosen parameters were sufficient for our study.

It is also important to emphasise that the computational cost of graph convolutions is proportional to the number of edges. As illustrated in Figure \ref{fig:subfig1}, the number of floating-point operations (FLOPs) required per event -- for all convolutions, averaged across the entire test dataset -- is influenced by the \textit{skip step} parameter. This analysis indicates that using connections with a~\textit{skip step} of 10 reduces FLOPs by a~factor of ten (3.568 vs 0.368 MFLOPs/ev), confirming a~linear relationship between computational demands and the number of generated edges. The role of our novel \textit{skip step} method is crucial for reducing the calculation complexity. Note that this is not specific to the context of FPGA implementations.

\subsubsection{\textbf{Model configurations}}

\begin{figure}[t]
    \centering
    \begin{subfigure}[b]{0.49\textwidth}
        \centering
        \includegraphics[width=\textwidth]{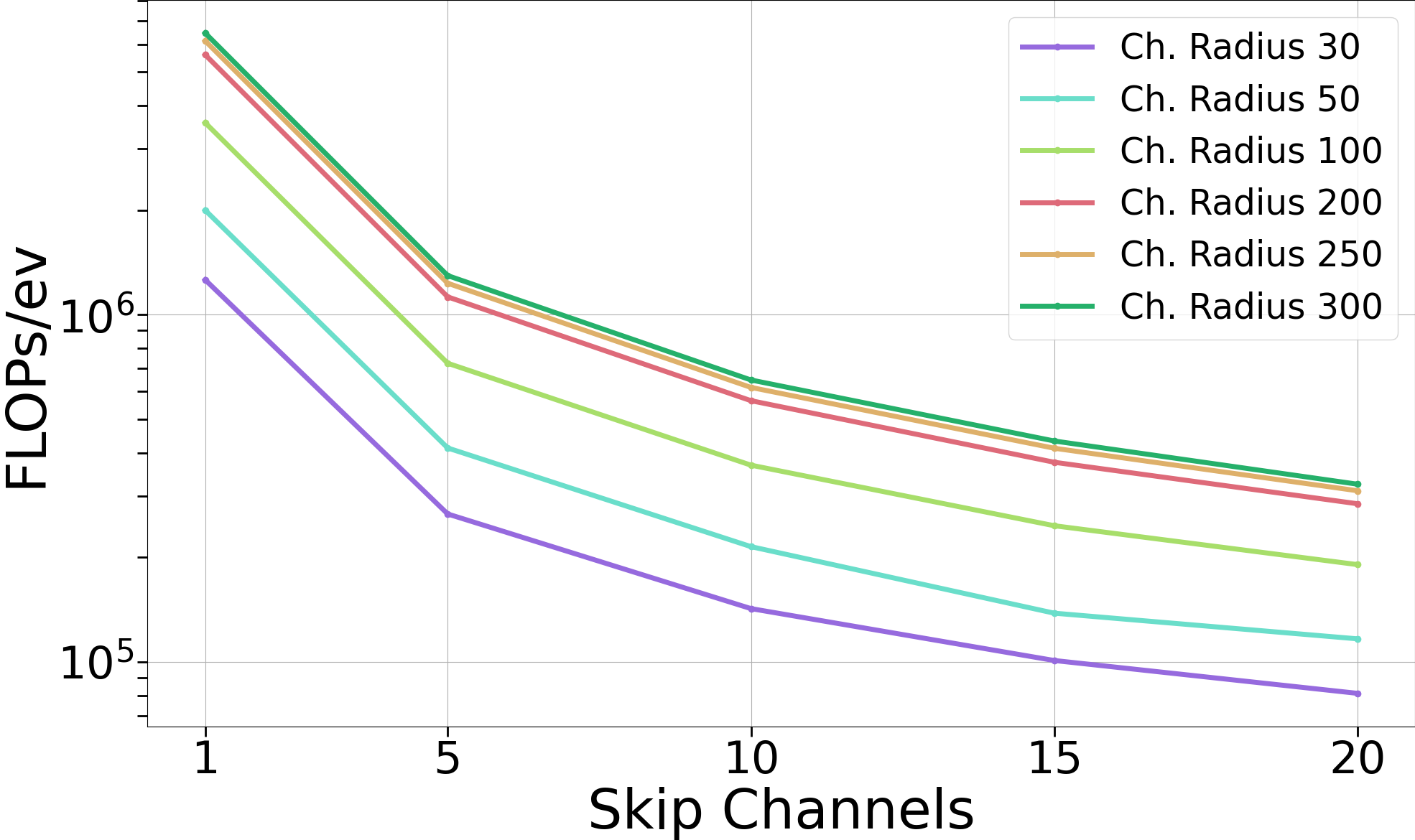}
        \caption{Average FLOPs per event for different graph generation configurations.}
        \label{fig:subfig1}
    \end{subfigure}
    \hfill
    \begin{subfigure}[b]{0.49\textwidth}
        \centering
        \includegraphics[width=\textwidth]{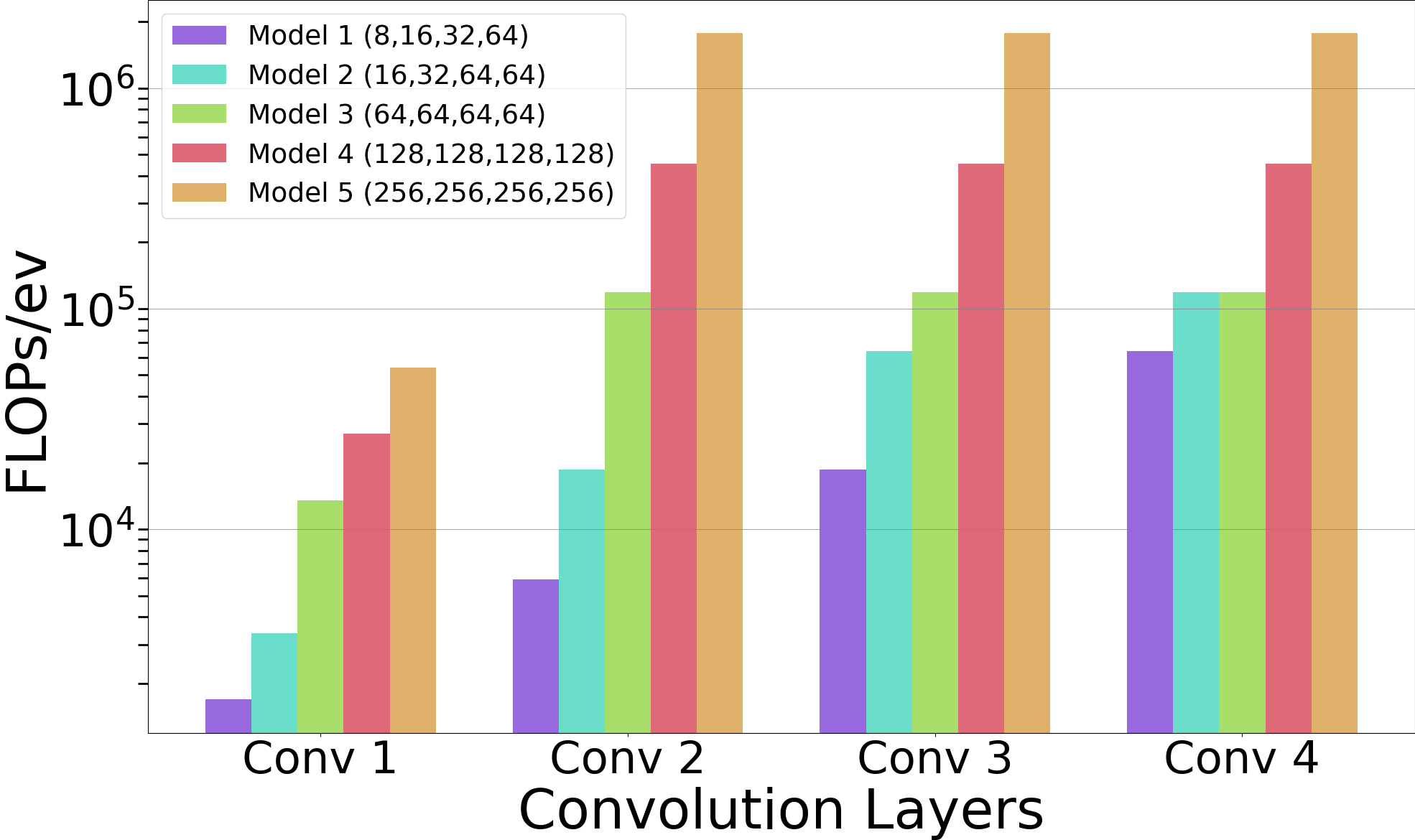}
        \caption{Average FLOPs per event for model configurations (outputs per convolution).}
        \label{fig:subfig2}
    \end{subfigure}
    \caption{Model complexity analysis in term of FLOPs. The figures illustrate not only the significant impact of model size on computational complexity, but also the parameters of the graph generator.}
    \Description{The figure shows model complexity analysis in terms of FLOPs. (a) The left plot presents average FLOPs per event for different graph generation configurations, showing how FLOPs decrease with increasing skip channels and vary with channel radius. (b) The right plot compares average FLOPs per event across convolution layers for different model sizes, highlighting the impact of convolution depth and output size on computational cost.}
    \label{fig:two-plots}
\end{figure}

In this subsection, we examine how varying the dimensions of the convolutional layers affects model performance. As a~reference, we used two model configurations described in \cite{lars}: small -- with all layer dimensions set to 64 and large -- with all layer dimensions set to 256. To explore a~broader range of options, we study three additional models: one with all layer sizes set to 128, and two variants with increasing layer size. The results, including layer dimensions, total parameter counts, and classification accuracy on both the SHD and SSC datasets, are summarised in Table \ref{tab:model_size}.

The results demonstrate that increasing the number of parameters improves classification accuracy on both datasets, highlighting the scalability of our method. Importantly, compared to the baseline (achieving 90.0\% and 94.3\% accuracy on SHD for the small and large models), the equivalent models based on our proposed approach (minor differences in parameter number are due to BatchNorm layers) achieve accuracies of 92.74\% and 94.63\%, respectively. 

\changed{A similar trend is observed for the SSC dataset. For SSC-35, accuracy increases steadily from 66.97\% for the smallest configuration to 71.02\% for the largest one, and for SSC-11 it improves from 78.93\% to 84.73\%. This improvement demonstrates that our method generalises well across tasks, consistently benefiting from higher model capacity. Furthermore, the performance gap between the floating-point and quantised implementations remains small (below 0.3\% in most cases), confirming that the proposed solutions not only facilitate efficient hardware implementation but also preserve accuracy after quantisation.}
\begin{table}[t]
\centering
\caption{Layer configurations (number of output features) with number of parameters and top-1 accuracy. Increasing the number of parameters improves classification results, demonstrating the scalability of our approach.}
\label{tab:model_size}
\resizebox{0.95\textwidth}{!}{%
\begin{tabular}{c c c c c c c c c c c c c}
\toprule
\multirow{2}{*}{Name} & 
\multirow{2}{*}{Conv1} & 
\multirow{2}{*}{Conv2} & 
\multirow{2}{*}{Conv3} & 
\multirow{2}{*}{Conv4} & 
\multirow{2}{*}{FC} & 
\multirow{2}{*}{Params [k]} & 
\multicolumn{2}{c}{SHD top-1 acc. [\%]} & 
\multicolumn{2}{c}{SSC-11 top-1 acc. [\%]} &
\multicolumn{2}{c}{SSC-35 top-1 acc. [\%]} \\ 
\cmidrule(lr){8-9} \cmidrule(lr){10-11} \cmidrule(lr){12-13}
 & & & & & & & float & quantised & float & quantised & float & quantised \\ 
\midrule
  tiny & 8  &  16 &  32 &  64 &  64 &   8.6   & 89.22 & 88.78 & 78.93 & 78.36 & 66.97 & 66.31 \\
 small & 16  &  32 &  64 &  64 &  64 &  12.9   & 90.54 & 90.98 & 80.63 & 80.16 & 67.49 & 67.21 \\
 base & 64  &  64 &  64 &  64 &  64 &  18.9   & 92.74 & 92.30 & 81.89 &  81.45 & 68.17 & 67.92 \\
 big & 128  & 128 & 128 & 128 & 128 &  70.5   & 93.93 & 93.31 & 83.11 & 82.79 & 69.54 & 69.36 \\
 large & 256  & 256 & 256 & 256 & 256 & 272.0   & 94.64 & 94.45 & 84.73 & 84.32 & 71.02 & 70.91 \\
\bottomrule
\end{tabular}%
}
\end{table}


Similar to the number of edges, another key factor influencing model complexity is the dimensionality of event feature vectors processed by the network. To address this, we conducted an analysis of the average number of FLOPs per event for each convolutional layer, depending on its configuration. The results are presented in Figure \ref{fig:subfig2}. They show that the computational complexity of the first convolutional layer increases linearly with its dimension due to the fixed size of input data (13.516 vs 27.032 kFLOPs/ev for Model 3 vs Model 4), while for subsequent layers in models with uniform dimensions, complexity grows quadratically (118.269 vs 452.801 kFLOPs/ev for Conv 2). Consequently, the total FLOPs for the evaluated models are 0.091, 0.204, 0.368, 1.385, and 5.366 MFLOPs/ev, respectively. These results underline the importance of careful selection of network parameters to optimise memory and computational costs.

\subsubsection{\textbf{Graph pooling}}
\label{sec:graph_pooling}

\begin{table*}[t]
\centering
\caption{Comparison of different operations in graph pooling for the SHD dataset:}
\label{tab:poolings}
\begin{subtable}[t]{0.45\linewidth}
\centering
\caption{classification task}
\label{tab:ablation_pooling_classification}
\resizebox{0.575\linewidth}{!}{%
\begin{tabular}{c c}
\toprule
Operation & Acc [\%] \\ 
\midrule
max       & 85.57 \\
\textbf{average} & \textbf{92.74} \\
add       & 87.81 \\
\bottomrule
\end{tabular}%
}
\end{subtable}%
\hfill
\begin{subtable}[t]{0.45\linewidth}
\centering
\caption{KWS task}
\label{tab:ablation_pooling_kws}
\resizebox{0.95\linewidth}{!}{%
\begin{tabular}{c c c}
\toprule
Operation & $Acc_K$ [\%] & $Acc_{K,\Delta}$ [\%] \\ 
\midrule
\textbf{max} & \textbf{90.11} & \textbf{87.19} \\
average      & 89.71          & 86.62 \\
add          & 87.32          & 83.26 \\
\bottomrule
\end{tabular}%
}
\end{subtable}
\end{table*}

Tables \ref{tab:ablation_pooling_classification} and \ref{tab:ablation_pooling_kws} present a~comparison of the applied operators in graph pooling for the classification task and the keyword spotting task. As described in Section \ref{sec:proposed_system}, in the classification task the aggregation of information from events processed by the feature extraction module was performed over the entire sample (on average 750 ms, up to a~maximum of 1 s). In contrast, for the KWS task, the aggregation was performed within shorter time windows of $\Delta t = 10$ ms.

As a~result, in the classification task, the average operator achieved the best performance, while the max and add operators yielded results lower by 7 pps and 5 pps, respectively. The max operator, when applied to such a~large number of events, proved to be sensitive to outliers. In the case of KWS, splitting the signal into shorter segments reduced the sensitivity to noise, which made the max operator slightly more effective (by 0.4 pps) compared to average. In both tasks, the add operator negatively impacted performance due to the accumulation of large values into a~single feature vector.



\subsection{Evaluation of the classification model}
\label{sec:sota}

\changed{Since the tasks of classification and keyword spotting, as well as their corresponding network models, differ substantially, we evaluate the proposed method for each task separately.
In this section, we present the results obtained with the classification model and compare them against related approaches evaluated on the same datasets.}

\begin{table}[t]
\centering
\caption{Comparison of accuracy and model size on the SHD and SSC datasets for the classification task. The symbol | indicates that only two models are present, while ``–'' denotes several intermediate values.}
\label{tab:big_table}
\resizebox{\textwidth}{!}{%
\begin{tabular}{@{}lllccccc@{}}
\toprule
\textbf{Dataset}               & \textbf{Author(s)}           & \textbf{Method}                             & \textbf{Recurrent} & \textbf{Hardware} & \textbf{\#Params}    & \textbf{top-1 acc. [\%]}    & \textbf{Year} \\ \midrule
\multirow{20}{*}{\textbf{SHD}} & Cramer et al. \cite{cramer2020}      & Recurrent SNN                      & \checkmarks         & \xmark        & -           & 83.2        & 2020 \\
                      & Yin et al. \cite{Yin2021}         & Adaptive SRNN                      & \checkmarks         & \xmark        & -           & 90.4        & 2021 \\
                      & Yao et al. \cite{yao2021}         & TA-SNN                             & \xmark         & \xmark        & -           & 91.1        & 2021 \\
                      & Yu et al. \cite{yu2022}          & STSC-SNN                           & \xmark         & \xmark        & 2.1M        & 92.4        & 2022 \\
                      & Dampfhoffer et al. \cite{dampfhoffer2022} & Cuba-LIF                           & \checkmarks         & \xmark        & 139k        & 87.8        & 2022 \\
                      & Bittar et al. \cite{bittar2022}    & Recurrent SNN                      & \checkmarks         & \xmark        & 3.9M        & 94.6        & 2022 \\
                      & Rossbroich et al. \cite{rossbroich2022}  & Convolutional RSNN                 & \checkmarks         & \xmark        & 209k        & 83.5        & 2022 \\
                      & Nowotny et al. \cite{nowotny2025loss}     & EventProp-GeNN                     & \checkmarks         & \xmark        & -           & 84.8        & 2022 \\
                      & Hammouamri et al. \cite{hammouamri2023}  & DCLS-Delays                        & \xmark         & \xmark        & 200k        & 95.1        & 2023 \\
                      & Sun et al. \cite{sun2023}         & DL128-SNN-Dloss                    & \xmark         & \xmark        & 140k        & 92.6        & 2023 \\
                      & Rafeldt et al. \cite{lars}     & Spectro-temporal graph             & \xmark         & \xmark        & 17.1 | 217k & 90.0 | 94.3 & 2024 \\
                      & D’Agostino et al. \cite{dagostino2024}  & FF SNN with dendritic delays       & \xmark         & \xmark        & 224 k       & 87.6        & 2024 \\
                      & Malettira et al. \cite{malettira2024}   & Temporal Skips with delay learning & \xmark         & \xmark        & 1.3M        & 94.7        & 2024 \\
                      & Carpegna et al. \cite{carpegna2024spiker}    & RSNN on FPGA                       & \checkmarks         & \checkmarks        & -           & 72.9        & 2024 \\
                      & Matinizadeh et al. \cite{quantisenc2024} & Fully configurable FPGA SNN        & \checkmarks         & \checkmarks        & -           & 87.8        & 2024 \\
                      & Schone et al. \cite{schone2024} & Event-SSM        & \checkmarks         & \xmark        & 0.4M           & 95.9        & 2024 \\ 
                      & Baronig et al. \cite{baronig2025} & SE-adLIF        & \checkmarks         & \xmark        & 450k           & 95.8        & 2025 \\
                      & Sun et al. \cite{sun2025} & PfA SNN       & \xmark         & \xmark        & 0.2M           & 96.3        & 2025 \\
                      & Huber et al. \cite{huber2024scaling}       & S5-RF                              & \checkmarks         & \xmark        & 214k        & 91.9        & 2025 \\
                      & \textbf{Ours}                & \textbf{Spectro-temporal graph on SoC FPGA} & \xmark         & \checkmarks        & \textbf{8.6 - 272k}  & \textbf{88.8 - 94.5} & \textbf{2025} \\ \midrule
\multirow{13}{*}{\textbf{SSC}} & Cramer et al. \cite{cramer2020}      & Recurrent SNN                      & \checkmarks         & \xmark        & -           & 50.9        & 2020 \\
                      & Perez-Nieves et al. \cite{perez2021} & Heter. RSNN                        & \checkmarks         & \xmark        & -           & 57.3        & 2021 \\
                      & Yin et al. \cite{Yin2021}         & Adaptive SRNN                      & \checkmarks         & \xmark        & -           & 57.3        & 2021 \\
                      & Dampfhoffer et al. \cite{dampfhoffer2022} & SpikGRU                            & \checkmarks         & \xmark        & 280k        & 77.0        & 2022 \\
                      & Bittar et al. \cite{bittar2022}      & Recurrent SNN                      & \checkmarks         & \xmark        & 3.9M        & 77.4        & 2022 \\
                      & Sadovsky et al. \cite{Sadovsky2023}    & SNN-CNN                            & \xmark         & \xmark        & -           & 72.0        & 2023 \\
                      & Hammouamri et al. \cite{hammouamri2023}  & DCLS-Delays                        & \checkmarks         & \xmark        & 0.7 - 2.5M  & 79.8 - 80.7 & 2023 \\
                      & Malettira et al. \cite{malettira2024}   & Temporal Skips with delay learning & \xmark         & \xmark        & 1.4M        & 80.2        & 2024 \\
                      & Schone et al. \cite{schone2024} & Event-SSM        & \checkmarks         & \xmark        & 0.1 | 0.6M           & 85.3 | 88.4        & 2024 \\    
                      & Baronig et al. \cite{baronig2025} & SE-adLIF        & \checkmarks         & \xmark        & 1.6M           & 80.4        & 2025 \\
                      & Sun et al. \cite{sun2025} & PfA SNN        & \xmark         & \xmark        & 0.1 | 0.7M           & 77.4 | 80.2       & 2025 \\
                      & Huber et al. \cite{huber2024scaling}       & S5-RF                              & \checkmarks         & \xmark        & 1.8M        & 78.8        & 2025 \\
                        & \textbf{Ours - SSC-11}                & \textbf{Spectro-temporal graph on SoC FPGA} & \xmark         & \checkmarks        &        \textbf{8.6 - 272k}     &   \textbf{78.4 - 84.3}          & \textbf{2025} \\
                      
                      & \textbf{Ours - SSC-35}                & \textbf{Spectro-temporal graph on SoC FPGA} & \xmark         & \checkmarks        &        \textbf{8.6 - 272k}     &   \textbf{66.3 - 70.9}          & \textbf{2025} \\ \bottomrule
\end{tabular}%
}
\end{table}

\subsubsection{\textbf{Comparison with the state-of-the-art}}
\label{sec:sota-class}

In order to position our work relative to the state-of-the-art, we first compare accuracy and parameter count of our event-graph neural network software model to other event-based AI approaches on the SHD and SSC datasets. Next, we compare our hardware implementation with two FPGA-based works of spiking neural networks, with consideration of power consumption, latency, and resource utilisation.

The results in Table \ref{tab:big_table} show that our models are among the smallest in the state-of-the-art while retaining competitive accuracy. On the SHD dataset, our base model requires only 18.9k parameters, being surpassed in compactness only by the small model from \cite{lars}, which inspired our baseline design, and by certain optimised variants of our own network (see Table \ref{tab:model_size}). Despite its small size, this model achieves a~test accuracy of 92.31\%, which is comparable to the results of \cite{sun2023} and \cite{yu2022}, while using 86.5\% and over 99\% fewer parameters, respectively. Our largest configuration achieves an accuracy of 94.45\%, which is outperformed by six competing models. However, two of these \cite{bittar2022,malettira2024} require millions of parameters, while two others \cite{schone2024,baronig2025} require more than 400k parameters, compared to only 272k in our case. Notably, the only solution surpassing ours by a~clear margin in both size and accuracy is \cite{hammouamri2023}, which achieves 95.1\% at 200k parameters. Importantly, our event-graph approach underwent extensive hardware-aware modifications and quantisation, which inevitably constrain the achievable peak accuracy. The high performance achieved under such constraints demonstrates the suitability of graph neural networks for event-data processing.

\changed{When extending the evaluation to the SSC dataset, we observe a~consistent pattern. Our smallest SSC model achieves 66.3\% top-1 accuracy, scaling up to 70.9\% for the largest configuration for SSC-35. On the other hand, for SSC-11, our models achieves from 78.4\% to 84.3\%. While this result falls short of the very best-performing software-only models (e.g. 88.4\% by \cite{schone2024} in case of SSC-35), it is important to note that all prior work on SSC remains purely in the software domain. To the best of our knowledge, no hardware implementation of SSC classification has been reported in the literature, highlighting a~significant gap in the field. By providing the first hardware-accelerated results on the SSC, we establish an important baseline for future work, showing that efficient event-graph architectures can be deployed under resource and quantisation constraints while still maintaining competitive performance.
}
\subsubsection{\textbf{Comparison with other hardware implementations}}
\label{sec:sota-class-hw}

\changed{A~direct comparison of our hardware implementation with spiking neural network (SNN)–based approaches from the literature is provided in Table~\ref{tab:comparison}. For this analysis, we selected the base model -- configured with a~graph convolution and head feature dimension of 64 -- implemented on the Zynq UltraScale+ ZCU104 board. This configuration was chosen as it provides a~favourable trade-off between accuracy and resource utilisation, offering relatively high accuracy while maintaining a~compact design.
In Section~\ref{sec:hw_scalablity}, we further report evaluation metrics for alternative model configurations to illustrate the scalability and flexibility of the proposed approach.}

\begin{table}[t]
\centering
\caption{Comparison of different variants of our architecture with state-of-the-art hardware implementations. The values in parentheses take into account the PS part (classification head) -- apart from our last model (PL-head), which was fully realised in the programmable logic part of the SoC FPGA.
The values marked with ``-'' were not mentioned in the respective papers.
The resource utilisation for \cite{quantisenc2024} was reported only in terms of percentages; thus, the absolute values were estimated based on the maximum resources available on the ZCU104 board.}
\resizebox{1.0\textwidth}{!}{%
\begin{tabular}{@{}lcccccccccc@{}}
\toprule
\textbf{Model}            & \textbf{Device}  & \textbf{Logic cells}  &  \textbf{LUT} &  \textbf{FF} &  \textbf{BRAM} &  \textbf{DSP} &  \textbf{Freq. [MHz]} &  \textbf{Latency [$\mu$s]} & \textbf{Power [W]} &  \textbf{Acc. [\%]} \\\midrule
Matinizadeh \cite{quantisenc2024} & Zynq ZCU104 & -  & 149,760 & 92,160 & 75 & - & 100  & - & 1.629$^1$ & 87.80 \\ 
Carpegna \cite{carpegna2024spiker} & Zynq Z7-20 & 18,268 & - & - & 51 & - & 100  & 540 & 0.43 & 72.99  \\
\midrule
Our (base) & Zynq ZCU104 & -  & 81,567 & 47,699 & 70 & 318 & 200  & 8.07 (179) & 0.99 (3.73) & 92.30  \\
Our (tiny) & Zynq ZCU104 & -  & 34,474 & 23,713 & 28 & 106 & 200  & 4.01 (175) & 0.78 (3.52) & 88.78 \\
Our (2VecMuls) & Zynq ZCU104 & -  & 41,925 & 30,198 & 70 & 164 & 200  & 15.2 (186)  & 0.89 (3.63) & 92.30 \\
Our (PL-head) & Zynq ZCU104 & - & 91,630  & 51,492 & 84.5 & 330 & 200  & 8.45 & 1.08 & 92.12 \\\bottomrule
\end{tabular}}
\label{tab:comparison}
\raggedright
\footnotesize{$^1$only the ``peak power'' was reported in the article.}
\end{table}

\changed{The proposed approaches for FPGA-implemented event-audio data classification, can be compared in terms of:
\begin{enumerate}

\item \textbf{Accuracy.} With our approach, we established a~new state-of-the-art accuracy for an FPGA solution applied to the SHD benchmark by a~considerable 4.5\% margin.

\item \textbf{Utilisation.} \changed{The detailed resource utilisation presented in Table~\ref{tab:comparison} demonstrates the applicability of our approach to relatively small embedded SoC FPGA platforms. To achieve a~more balanced use of available logic resources compared to \cite{quantisenc2024}, we implemented part of the multiplication operations using DSP modules, thereby mitigating potential routing delays. The hardware modules are parametrised such that each multiplication can be flexibly mapped either to DSPs or logic resources.
Although graph convolutional neural networks are often criticised for their memory-intensive inference \cite{wang2021empirical} -- stemming from the need to store and access features of neighbouring nodes -- our optimised memory management, combined with the proposed enhancements to graph generation, resulted in memory utilisation comparable to that of spiking neural network–based systems.}

\item \textbf{Latency.} An important advantage of our solution is its asynchronous event-by-event processing and the intrinsic ability of event-graphs to exploit temporal sparsity for rapid and efficient calculations.
However, this makes it difficult to compare it with other solutions in terms of latency. 
\changed{Since each event is processed immediately upon registration and the hardware module does not rely on prior aggregation, we define latency in this study as the time elapsed between the arrival of the last event in the processed sample and the generation of the corresponding prediction. Using this methodology, we measured the per-event latency for graph generation and feature extraction to be 8 µs at a~200 MHz clock frequency, with a~throughput of 555 kEPS.}
After taking into account the PS-PL communication and the determination of the value of classification results by the network's head in the processing part of the heterogeneous system, we obtain the predicted class 179 $\mu$s after the occurrence of the last event.
In \cite{carpegna2024spiker} the reported latency of 0.54 ms assumes the processing of the entire data sample after its encoding in 100 timesteps. 
Our solution, which requires no prior data aggregation, maintains high throughput and low latency for edge applications by exploiting data's temporal sparsity.

\item \textbf{Power.} 
As our system is designed to exploit the inherent sparsity of event data, its dynamic power consumption strongly depends on the amount of activity registered by the AC sensor. To obtain an accurate estimate of average power consumption, we followed the methodology of \cite{quantisenc2024}: we use simulations to extract net toggle rates, which are then supplied to the Vivado Power Analyzer to compute average dynamic power. For this purpose, we generated a~representative custom data sample with events occurring every 50 $\mu$s (consistent with an average event rate of 20 kEPS for the SHD dataset). Event frequencies (channels) were selected to create a~worst-case scenario in terms of the number of graph edges.
The power estimation provided by the Vivado software for the PL part of the system (base model) is 0.99 W (0.396 W dynamic and 0.594 W static). With the PS part included energy usage increases to 3.73 W (3.25 W dynamic and 0.7 W static) for the entire base architecture.

While the average power consumption of our system -- when relying on the processing system to generate the final prediction -- is higher than that reported in \cite{carpegna2024spiker} and \cite{quantisenc2024}, the substantially higher accuracy and smaller latency achieved with our approach justifies this additional cost. 
Moreover, by integrating the network head directly into the programmable logic, we obtain far more competitive results, as demonstrated in Section~\ref{sec:hw_scalablity}.

\end{enumerate}

}
 
\subsubsection{\textbf{Hardware module scalability}}
\label{sec:hw_scalablity}

\changed{

A~key advantage of the proposed method lies in its scalability and flexibility. In this section, we present a~series of experiments conducted on the time-series classification model to evaluate modified architectures against the baseline. The resulting metrics are summarised in Table~\ref{tab:comparison}.

\subsubsection*{\textbf{Evaluation of the tiny model}}

To demonstrate the scalability of our hardware module, we implemented the tiny configuration, which represents the smallest model described in Table~\ref{tab:model_size}. Reducing the output dimensions of the graph convolution modules leads to a~substantial decrease in logic, DSP, and memory utilisation. This outcome is expected, as the smaller model requires fewer multiplications and stores fewer features. An additional benefit is a~significantly reduced latency, approximately 50\% of that observed in the baseline implementation for graph generation and feature extraction.
The power consumption estimated for this configuration is also the smallest, as expected (just 0.186 W dynamic, 0.593 W static).

It is noteworthy that the tiny model achieved 0.8\% higher accuracy while utilising only 23\% of LUTs and 37\% of BRAMs compared to \cite{quantisenc2024} -- see Table \ref{tab:comparison}.

\subsubsection*{\textbf{Decreased number of parallel multiplications}}

For an additional ablation study we implemented the resource-optimised variant of the base model for decreased number of parallel multiplications (from 4 per convolution to 2) -- \textbf{2VecMuls model} in Table~\ref{tab:comparison}.
With this modification, we were able to achieve 49\% decrease in both DSP and LUT utilisation and 35\% in FFs with a~simultaneous increase of latency for a~single event (in the PL) to 15.2 $\mu$s (theoretical throughput of 277 kEPS).
This experiment confirms that the systems latency and resource utilisation (and thus power consumption) is highly correlated -- the choice of the final solution should be based on the requirements of the specific task and the size of the target platform.

\subsubsection*{\textbf{Classification head integrated into programmable logic}}

As indicated in Section \ref{sec:classhead}, while in the baseline implementation the classification head was developed for the processing system of the heterogenous platform, it is possible to implement entire network pipeline for the programmable logic as end-to-end system. 
For this purpose, we implemented two MLP layers with ReLU activation, computed using two vector multiplication modules (processing two rows in parallel). With this modification, logic resource utilisation increased by 12\%, flip-flops by 8\%, BRAMs by 21\%, and DSPs by 4\%. However, in this configuration, the use of the processing system portion of the heterogeneous device is no longer necessary, resulting in a~significant reduction in both latency (time between receiving the last event and generating the prediction is 8.45 $\mu$s) and power consumption (estimated at 1.08 W, comprising 0.486 W dynamic and 0.595 W static power). 
As the weights in the network's head have been quantised for hardware implementation, the network's accuracy is 0.18 pps smaller.

The particular configuration of the proposed hardware system can be chosen based on specific application's requirements and resource availability on the target platform.
}

\subsection{Evaluation of the KWS model}
\label{sec:sota-kws}

\changed{In this section, we present the evaluation of an end-to-end implementation of the KWS module on the programmable logic of a~heterogeneous FPGA platform, specifically the Zynq UltraScale+ ZCU104 board. For this experiment, all layers (including the output head) were quantised to 8-bit integer values. The output feature size of each graph convolution layer, as well as the STEM and GRU modules, was fixed at 72 in order to efficiently utilise the BRAM blocks and their 36-bit word size.

The hardware module was first verified through Vivado simulation to ensure compatibility with the software model and subsequently implemented on the target platform. 
For the purposes of this evaluation, the clock frequency was set to 200 MHz.
It should be noted, however, that the KWS model design does not restrict the user from selecting lower clock frequencies, thereby enabling reduced power consumption at the cost of increased latency.}

\begin{table}[t]
\centering
\caption{Results of the floating-point (FP32) and quantised (8-bit) models for the keyword spotting task on the SHD and SSC datasets.}
\label{tab:SW_kws}
\resizebox{.8\textwidth}{!}{%
\begin{tabular}{l c c c c c}
\toprule
\multirow{2}{*}{Dataset} & \multirow{2}{*}{\#Params} & \multicolumn{2}{c}{Float} & \multicolumn{2}{c}{Quantised} \\
\cmidrule(lr){3-4} \cmidrule(lr){5-6}
 &  & $Acc_K$ [\%] & $Acc_{K, \Delta}$ [\%] & $Acc_K$ [\%] & $Acc_{K, \Delta}$ [\%] \\
\midrule
SHD    & 60.34k & 90.11 & 87.19 & 88.69 & 85.91 \\
SSC-11 & 59.77k & 76.21 & 73.66 & 73.49 & 71.30 \\
SSC-35 & 61.52k & 66.80 & 64.45 & 64.74 & 63.12 \\
\bottomrule
\end{tabular}%
}
\end{table}

\changed{The results of software-based model training and corresponding model sizes are presented in Table~\ref{tab:SW_kws}. The average model size is approximately 60k parameters, with minor variations due to the different numbers of output classes. On the SHD dataset, accuracy decreases from 92.74\% for the base classification task to 90.11\% in the KWS task (FP32), reflecting the higher complexity of the latter, where classification relies on the maximum confidence score across the entire sample. Furthermore, when comparing top-1 accuracy with keyword detection at the generated offset, the accuracy is approximately 87.19\%, which is a~satisfactory outcome for such a~challenging task.

Similarly, for the SSC dataset, both variants (11 and 35 classes) demonstrate that the difference between $Acc_K$ and $Acc_{K, \Delta}$ is around 2.5\%. For all datasets, the models successfully detect the end of the keyword (within a~$\pm 1 \Delta t$ offset) in more than 95\% of cases. Additionally, reducing the number of classes in SSC increases accuracy by approximately 10\%. 
}

\changed{The evaluation metrics of the KWS model implemented in hardware are presented in Table~\ref{tab:HW-kws}. Our method achieves high accuracy (see Table~\ref{tab:SW_kws} -- quantised results) with a~low latency of only 10.53~$\mu$s for end-to-end KWS task with prediction generated every 10 ms. Latency is measured as the time between receiving the last event in a~time window and generating the prediction. The low resource utilisation further enables deployment on relatively small, embedded FPGA platforms.
Moreover, this architecture can be applied to continuous event streams, as the recurrent head naturally supports such processing.

Average power consumption was estimated using the Vivado Power Analyzer, based on net toggle rates extracted from a~representative data sample. The event rate was aligned with the average of 20~kEPS established for the SHD dataset, while frequencies were chosen to represent a~worst-case scenario in terms of the number of graph edges. The achieved average consumption of 1.184~W (0.588 dynamic, 0.595 static) underscores the applicability of our method to energy-constrained use cases, such as Internet-of-Things edge devices.
As indicated in Table~\ref{tab:HW-kws}, while the network head consumes a~considerable portion of resources (20\% of all LUTs and 33\% of all FFs used in the system), it accounts for only 13\% of the total power consumption. This is due to its activity for just 2.06 $\mu$s every 10~ms, being triggered only after the pooling module.

To the best of our knowledge, this work represents the first hardware implementation of a~KWS system that processes audio in the event-driven domain. As such, it can serve as a~benchmark for future research. }

\begin{table}[t]
\centering
\caption{Resource utilisation, latency and power consumption for KWS model as described in Section \ref{sec:sota-kws}. We report the metrics for entire architecture as well as graph generation + feature extraction + pooling (FE + pool) and GRU head separately. }
\resizebox{0.6\textwidth}{!}{%
\begin{tabular}{@{}lccc@{}}
\toprule
\textbf{}            & 
\textbf{FE + pool} & \textbf{GRU head} & \textbf{Total}\\\midrule
LUT                & 99,940 & 25,190 & 125,130  \\
FF                 & 55,182 & 27,190 & 82,372   \\
BRAM               & 58.5 & 17 & 75.5    \\
DSP                & 108 & 32 & 140     \\
Latency [$\mu$s]   & 8.48 & 2.05 & 10.53        \\
Average power [W]  & 1.03 & 0.15 & 1.18
\\\bottomrule
\end{tabular}}
\label{tab:HW-kws}
\end{table}

\subsection{Comparison with embedded GPU}
\label{sec:power}

\begin{figure*}[t]
  \centering
  \begin{subfigure}[t]{0.32\textwidth}
    \includegraphics[width=\linewidth]{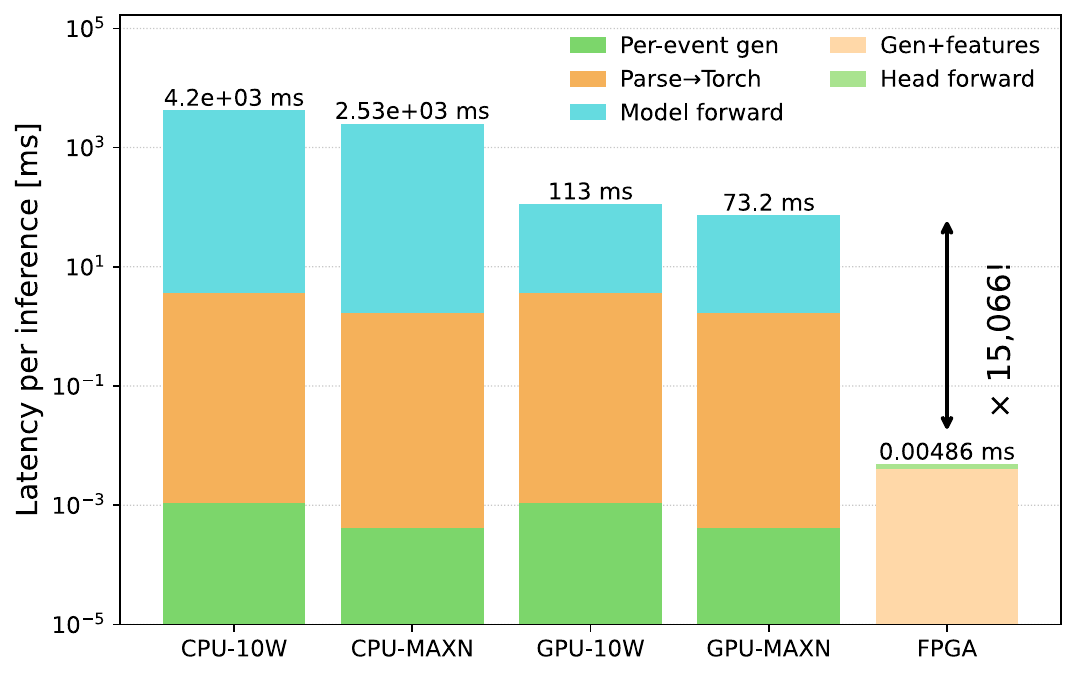} 
    \caption{Tiny model classification}
  \end{subfigure}
  \hfill
  \begin{subfigure}[t]{0.32\textwidth}
    \includegraphics[width=\linewidth]{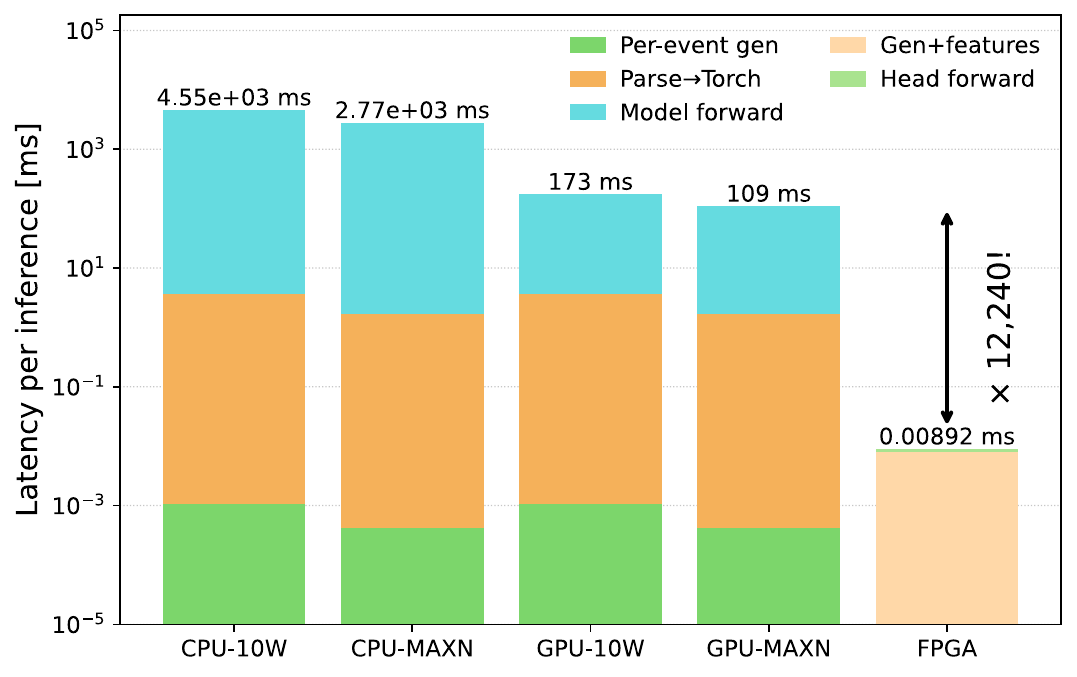}
    \caption{Base model classification}
  \end{subfigure}
  \hfill
  \begin{subfigure}[t]{0.32\textwidth}
    \includegraphics[width=\linewidth]{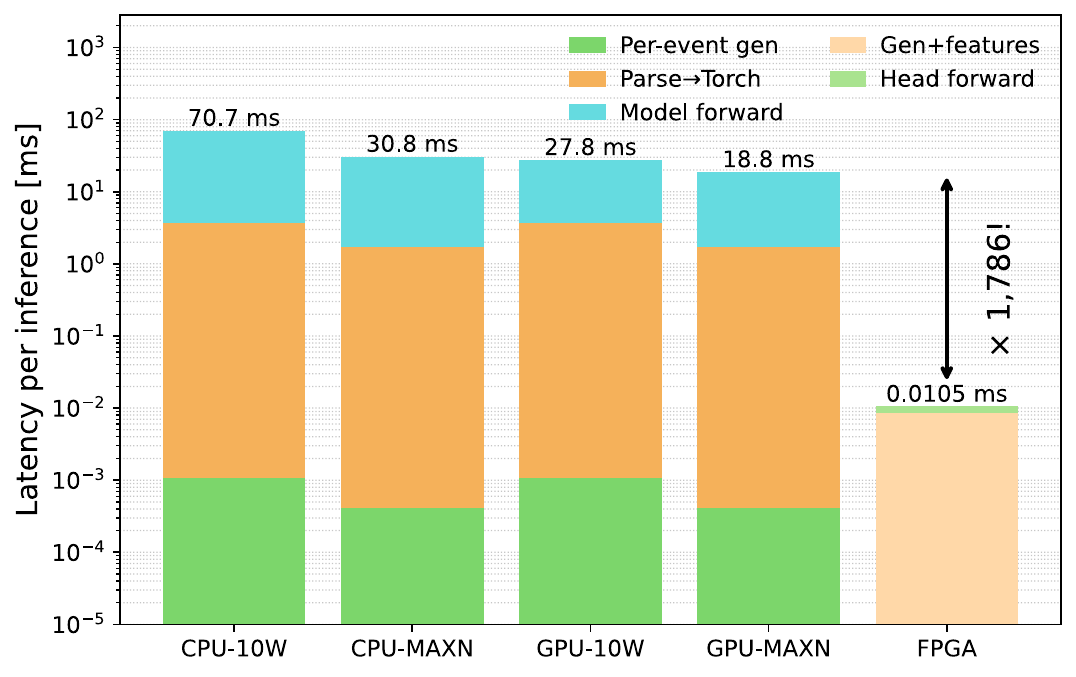}
    \caption{Keyword spotting}
  \end{subfigure}
  \caption{Latency comparison of tiny, base, and keyword spotting models on Jetson Orin NX (CPU/GPU) and ZCU104 FPGA.}
  \Description{The figure shows a~latency comparison of tiny, base, and keyword spotting models on Jetson Orin NX (CPU/GPU) and ZCU104 FPGA. (a) Tiny model classification, (b) base model classification, and (c) keyword spotting are evaluated. The stacked bars indicate contributions from pre-event generation, event processing, graph feature extraction, and model forward pass. Results highlight that the FPGA achieves significantly lower latency compared to CPU and GPU implementations.}
  \label{fig:latency}
\end{figure*}

\begin{figure*}[t]
  \centering
    \includegraphics[width=\linewidth]{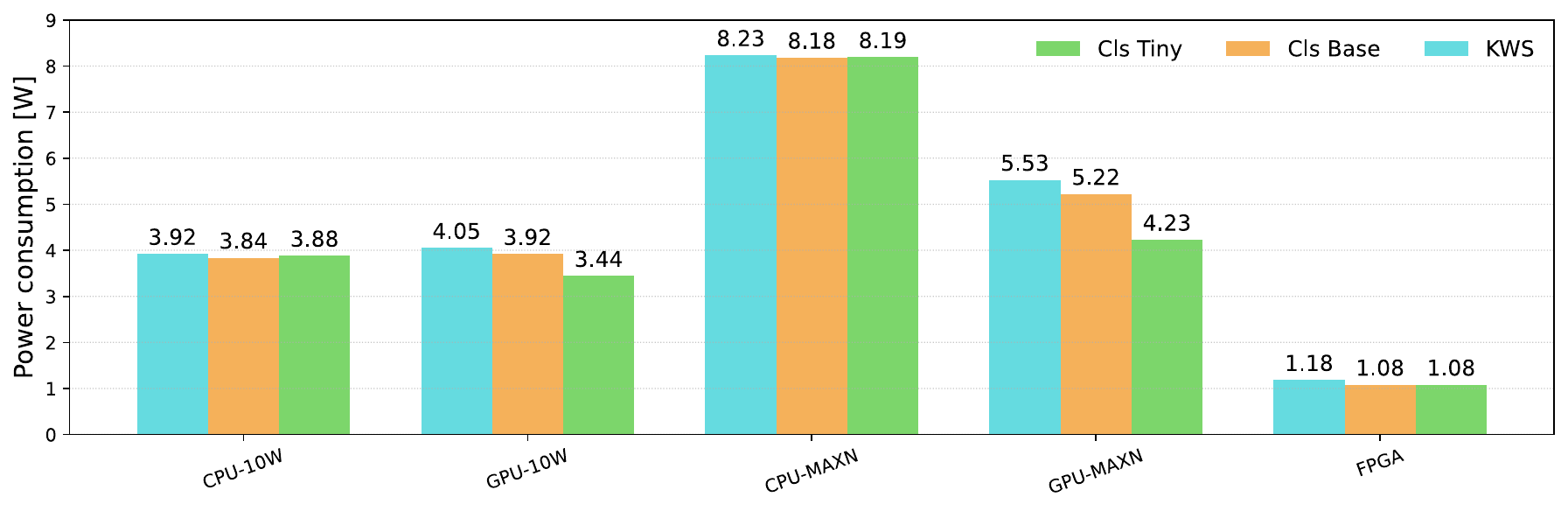} 
  \caption{Power consumption of tiny, base, and keyword spotting models on Jetson Orin NX (CPU/GPU) and ZCU104 FPGA.}
  \Description{The figure presents the power consumption of tiny, base, and keyword spotting models on Jetson Orin NX (CPU/GPU) and ZCU104 FPGA. CPU and GPU implementations show significantly higher power usage, especially in MAXN modes, while the FPGA achieves the lowest consumption (~1.1 W) across all model types, demonstrating its energy efficiency.}
  \label{fig:power}
\end{figure*}

\changed{To evaluate the temporal and energy efficiency of the proposed method in comparison with alternative embedded platforms, a~reference implementation was developed on the Jetson Orin NX platform. This system is equipped with an ARM Cortex-A78 processor and an NVIDIA Ampere GPU. To ensure comparability with the FPGA configuration, a~graph generator was implemented in C++ and executed on a~single CPU core in an event-by-event manner. The generated graph was subsequently processed by the model on both the GPU and CPU at two power modes: 10 W and MAXN.

In the Jetson setup, three latency-inducing stages can be distinguished: (i) per-event graph generation, (ii) parsing the generated graph into a~Torch tensor, and (iii) model inference. In contrast, on the FPGA platform, the graph generator, feature extractor and graph pooling operate in an event-by-event manner, after which only the head processes the pooled graph outputs. For each task, latency is measured after the arrival of the last event within the temporal window (1 s for classification and 10 ms for keyword spotting).

Figure~\ref{fig:latency} reports the latency of each stage (logarithmic scale on the y-axis). Since graph generation and parsing are executed on the CPU, their values are comparable in both CPU and GPU-based configurations. At the 10 W power level, graph generation requires on average 1 $\mu$s per event, whereas at MAXN this time is reduced to 0.42 $\mu$s. Parsing the graph into a~tensor takes 3.69 ms and 1.71 ms, respectively. Considering only the fastest Jetson configuration (GPU-MAXN), forward propagation for the tiny/base/KWS models requires 71.51 ms, 107.47 ms, and 96.16 ms, respectively. By contrast, our FPGA implementation achieves 4.01 $\mu$s, 8.07 $\mu$s, and 8.48 $\mu$s per event for graph generation and feature extraction, as well as 0.85 $\mu$s and 2.05 $\mu$s for classification and keyword spotting heads. These results correspond to speed-ups of more than $15$,$000\times$, $12$,$000\times$, and $1$,$700\times$ compared to Jetson.

Figure~\ref{fig:power} presents the power consumption results for each configuration. The measurements demonstrate that our hardware implementation on the ZCU104 consumes up to $4\times$ less power than the GPU-MAXN configuration and up to $8\times$ less than the CPU-MAXN, highlighting that our solution surpasses Jetson in both latency and power efficiency.}

\section{Summary}
\label{sec:summary}

In this work we present the first hardware implementation of an event-graph neural network for time-series audio classification and keyword spotting tasks. As our approach exploits the inherent sparsity of event-data to reduce the computational complexity and latency, it is highly promising for near-sensor AI processing at the edge.
We proposed adaptations and optimisations permitting an event-graph to be implemented efficiently in a~SoC FPGA. In particular, we presented the novel \textit{skip step} graph generation method with simplified and computationally efficient features. 


In spite of this, for time-series classification our quantised hardware-aware event-graph model achieved a~test accuracy extremely close to floating-point precision software models from the state-of-the-art while requiring almost two orders of magnitude fewer parameters. Crucially, our method outperformed all previous FPGA implementations of hardware-aware spiking neural networks on the same classification benchmark achieving improvements of 4.5\% and 19.3\% in accuracy. Relative to these works, the utilisation of FPGA resources and the latency were also reduced.
Our smallest model (tiny) achieves accuracy comparable to previous hardware state-of-the-art implementations while using only 23\% of the logic and 37\% of the memory resources. 
Moreover, our end-to-end implementation -- which includes graph generation, feature extraction, and the classification head -- supports real-time processing with a~latency of just 8.45 $\mu$s and an average power consumption of 1.08 W, achieving 4.32 pps higher accuracy with a~relatively similar model size compared to \cite{quantisenc2024}.

We also proposed a~separate architecture for a~keyword spotting task, that utilises graph convolutional neural networks for feature extraction and a~recurrent head for final prediction. 
Our end-to-end hardware implementation enables real-time KWS on continuous stream with prediction generated every 10 ms with the latency of just 10.53 $\mu$s and average power consumption estimated at 1.18~W. 

To ensure comparability, we evaluated our model on open-source datasets: SHD, SSC-11, and SSC-35, achieving tolerance-aware top-1 accuracy of 85.91\%, 71.30\%, and 63.12\%, respectively. Additionally, the model attained a~word-end detection accuracy exceeding 95\%, thereby enabling other researchers to directly assess and extend our results.

To the best of our knowledge, there are no prior works addressing keyword spotting on these datasets; thus, our results establish a~baseline for this task. Nevertheless, several strategies could further improve performance, such as data augmentation through random noise injection or random temporal offsets.

Additionally, we conducted a~comparison of our solution with the implementation on the Jetson Orin NX, which demonstrated that our hardware architecture achieves latency reductions of approximately $15$,$000\times$, $12$,$000\times$, and $1$,$700\times$ for the tiny, base, and keyword spotting models, respectively, while reducing energy consumption by more than $4\times$ compared to the GPU and $8\times$ compared to the CPU.

We have confirmed that the hardware implementation of graph neural networks applied to event data from artificial cochlea sensors is both highly efficient and capable of achieving superior task accuracy and latency compared to spiking neural network alternatives. 
In future work, we intend to integrate real AC sensor with our FPGA implementation of the graph neural network for a~fully end-to-end system. We also plan to implement NAS sensor in FPGA's resources and connect it with our hardware module to allow a~comparison of these two approaches. Finally, we intend to deploy the complete system on an ASIC to further reduce power consumption.


\section{Acknowledgments}
This work was supported by The Horizon Europe (dAIedge, grant 101120726), the ``Excellence initiative –- research university'' programme for the AGH University of Krakow, the Polish National Science Centre projects 2024/53/N/ST6/04254 and 2024/53/N/ST6/04331 and Polish high-performance computing infrastructure PLGrid (HPC Center: ACK Cyfronet AGH -- grant no. PLG/2023/016897).
\bibliographystyle{ACM-Reference-Format}
\bibliography{DAS-KWS}

@misc{jeziorek2024embedded,
      title={{Embedded Graph Convolutional Networks for Real-Time Event Data Processing on SoC FPGAs}}, 
      author={Kamil Jeziorek and Piotr Wzorek and Krzysztof Blachut and Andrea Pinna and Tomasz Kryjak},
      year={2024},
      eprint={2406.07318},
      archivePrefix={arXiv},
      primaryClass={cs.CV},
      url={https://arxiv.org/abs/2406.07318}, 
}

@inproceedings{li2021graph,
  title={Graph-based asynchronous event processing for rapid object recognition},
  author={Li, Yijin and Zhou, Han and Yang, Bangbang and Zhang, Ye and Cui, Zhaopeng and Bao, Hujun and Zhang, Guofeng},
  booktitle={Proceedings of the IEEE/CVF International Conference on Computer Vision},
  pages={934--943},
  year={2021},
  doi = {10.1109/ICCV48922.2021.00097}
}

@INPROCEEDINGS{dalgaty2023hugnet,
    author    = {Dalgaty, Thomas and Mesquida, Thomas and Joubert, Damien and Sironi, Amos and Vivet, Pascal and Posch, Christoph},
    title     = {HUGNet: Hemi-Spherical Update Graph Neural Network Applied to Low-Latency Event-Based Optical Flow},
    booktitle = {Proceedings of the IEEE/CVF Conference on Computer Vision and Pattern Recognition (CVPR) Workshops},
    month     = {June},
    year      = {2023},
    pages     = {3952-3961},
    doi =  {10.1109/CVPRW59228.2023.00411}
}

@inproceedings{jeziorek2023memory,
  title={Memory-efficient graph convolutional networks for object classification and detection with event cameras},
  author={Jeziorek, Kamil and Pinna, Andrea and Kryjak, Tomasz},
  booktitle={2023 Signal Processing: Algorithms, Architectures, Arrangements, and Applications (SPA)},
  pages={160--165},
  year={2023},
  organization={IEEE},
  doi = {10.23919/SPA59660.2023.10274464}
}

@ARTICLE{yang2024evgnn,

  author={Yang, Yufeng and Kneip, Adrian and Frenkel, Charlotte},
  journal={IEEE Transactions on Circuits and Systems for Artificial Intelligence}, 
  title={EvGNN: An Event-Driven Graph Neural Network Accelerator for Edge Vision}, 
  year={2025},
  volume={2},
  number={1},
  pages={37-50},
  doi={10.1109/TCASAI.2024.3520905}}

@inproceedings{mesquida2023g2n2,
  TITLE = {{G2N2: Lightweight event stream classification with GRU graph neural networks}},
  AUTHOR = {Mesquida, Thomas and Dampfhoffer, Manon and Dalgaty, Thomas and Vivet, Pascal and Sironi, Amos and Posch, Christoph},
  URL = {https://cea.hal.science/cea-04321175},
  BOOKTITLE = {{https://proceedings.bmvc2023.org/}},
  ADDRESS = {Aberdeen, United Kingdom},
  SERIES = {https://proceedings.bmvc2023.org/},
  PAGES = {660},
  YEAR = {2023},
  MONTH = Nov,
  HAL_ID = {cea-04321175},
  HAL_VERSION = {v1},
}

@inproceedings{dalgaty2023cnn,
  title={{The CNN vs. SNN Event-camera Dichotomy and Perspectives For Event-Graph Neural Networks}},
  author={Dalgaty, Thomas and Mesquida, Thomas and Joubert, Damien and Sironi, Amos and Soubeyrat, Cyrille and Vivet, Pascal and Posch, Christoph},
  booktitle={2023 Design, Automation \& Test in Europe Conference \& Exhibition (DATE)},
  pages={1--6},
  year={2023},
  organization={IEEE},
  doi = {10.23919/DATE56975.2023.10137023}
}

@article{wang2021empirical,
  title={Empirical analysis of performance bottlenecks in graph neural network training and inference with GPUs},
  author={Wang, Zhaokang and Wang, Yunpan and Yuan, Chunfeng and Gu, Rong and Huang, Yihua},
  journal={Neurocomputing},
  volume={446},
  pages={165--191},
  year={2021},
  publisher={Elsevier}
}

@ARTICLE{carpegna2024spiker,
author={Carpegna, Alessio and Savino, Alessandro and Carlo, Stefano Di},
journal={ IEEE Transactions on Emerging Topics in Computing },
title={{Spiker+: a framework for the generation of efficient Spiking Neural Networks FPGA accelerators for inference at the edge}},
year={2024},
volume={},
number={01},
ISSN={2168-6750},
pages={1-15},
doi={10.1109/TETC.2024.3511676},
publisher={IEEE Computer Society},
address={Los Alamitos, CA, USA},
month=dec}

@article{dalgaty2024mosaic,
  title={Mosaic: in-memory computing and routing for small-world spike-based neuromorphic systems},
  author={Dalgaty, Thomas and Moro, Filippo and Demira{\u{g}}, Yi{\u{g}}it and De Pra, Alessio and Indiveri, Giacomo and Vianello, Elisa and Payvand, Melika},
  journal={Nature Communications},
  volume={15},
  number={1},
  pages={142},
  year={2024},
  publisher={Nature Publishing Group UK London},
  doi = {10.1038/s41467-023-44365-x}  
}

@article{xu2023event,
  title={Event-driven spectrotemporal feature extraction and classification using a silicon cochlea model},
  author={Xu, Ying and Perera, Samalika and Bethi, Yeshwanth and Afshar, Saeed and van Schaik, Andr{\'e}},
  journal={Frontiers in Neuroscience},
  volume={17},
  pages={1125210},
  year={2023},
  publisher={Frontiers Media SA},
  doi = {10.3389/fnins.2023.1125210}
}

@inproceedings{kryjak2024event,
  title={Event-Based Vision on FPGAs-a Survey},
  author={Kryjak, Tomasz},
  booktitle={2024 27th Euromicro Conference on Digital System Design (DSD)},
  pages={541--550},
  year={2024},
  organization={IEEE},
  doi = {10.1109/DSD64264.2024.00078}
}

@inproceedings{basu2022spiking,
  title={Spiking neural network integrated circuits: A review of trends and future directions},
  author={Basu, Arindam and Deng, Lei and Frenkel, Charlotte and Zhang, Xueyong},
  booktitle={2022 IEEE Custom Integrated Circuits Conference (CICC)},
  pages={1--8},
  year={2022},
  organization={IEEE},
  doi={10.1109/CICC53496.2022.9772783}
}

@inproceedings{ortner2023online,
  title={Online spatio-temporal learning with target projection},
  author={Ortner, Thomas and Pes, Lorenzo and Gentinetta, Joris and Frenkel, Charlotte and Pantazi, Angeliki},
  booktitle={2023 IEEE 5th International Conference on Artificial Intelligence Circuits and Systems (AICAS)},
  pages={1--5},
  year={2023},
  organization={IEEE},
  doi = {10.1109/AICAS57966.2023.10168623}
}

@article{liu2013asynchronous,
  title={{Asynchronous Binaural Spatial Audition Sensor With 2 $\times$64$\times$4 Channel Output}},
  author={Liu, Shih-Chii and van Schaik, Andre and Minch, Bradley A and Delbruck, Tobi},
  journal={IEEE Transactions on Biomedical Circuits and Systems},
  volume={8},
  number={4},
  pages={453--464},
  year={2013},
  publisher={IEEE},
  doi = {10.1109/TBCAS.2013.2281834}
}

@article{gallego2020event,
  title={Event-based vision: A survey},
  author={Gallego, Guillermo and Delbr{\"u}ck, Tobi and Orchard, Garrick and Bartolozzi, Chiara and Taba, Brian and Censi, Andrea and Leutenegger, Stefan and Davison, Andrew J and Conradt, J{\"o}rg and Daniilidis, Kostas and others},
  journal={IEEE Transactions on Pattern Analysis and Machine Intelligence},
  volume={44},
  number={1},
  pages={154--180},
  year={2020},
  publisher={IEEE},
  doi = {10.1109/TPAMI.2020.3008413}
}

@article{dimarco2003implantable,
  title={Implantable cardioverter--defibrillators},
  author={DiMarco, John P},
  journal={New England Journal of Medicine},
  volume={349},
  number={19},
  pages={1836--1847},
  year={2003},
  publisher={Mass Medical Soc},
  doi = {10.1056/NEJMra035432 }
}

@article{saufi2020gearbox,
  title={Gearbox fault diagnosis using a deep learning model with limited data sample},
  author={Saufi, Syahril Ramadhan and Ahmad, Zair Asrar Bin and Leong, Mohd Salman and Lim, Meng Hee},
  journal={IEEE Transactions on Industrial Informatics},
  volume={16},
  number={10},
  pages={6263--6271},
  year={2020},
  publisher={IEEE},
  doi = {10.1109/TII.2020.2967822}
}

@article{ren2023deep,
  title={{Deep learning for time-series prediction in IIoT: progress, challenges, and prospects}},
  author={Ren, Lei and Jia, Zidi and Laili, Yuanjun and Huang, Di},
  journal={IEEE Transactions on Neural Networks and Learning Systems},
  year={2023},
  publisher={IEEE},
  doi = {10.1109/TNNLS.2023.3291371 }
}

@article{lim2021time,
  title={Time-series forecasting with deep learning: a survey},
  author={Lim, Bryan and Zohren, Stefan},
  journal={Philosophical Transactions of the Royal Society A},
  volume={379},
  number={2194},
  pages={20200209},
  year={2021},
  publisher={The Royal Society Publishing},
  doi = {10.1098/rsta.2020.0209}
}

@INPROCEEDINGS{POINTNET,
  author={Charles, R. Qi and Su, Hao and Kaichun, Mo and Guibas, Leonidas J.},
  booktitle={2017 IEEE Conference on Computer Vision and Pattern Recognition (CVPR)}, 
  title={PointNet: Deep Learning on Point Sets for 3D Classification and Segmentation}, 
  year={2017},
  volume={},
  number={},
  pages={77-85},
  doi={10.1109/CVPR.2017.16}}

@ARTICLE{cramer2020,
  author={Cramer, Benjamin and Stradmann, Yannik and Schemmel, Johannes and Zenke, Friedemann},
  journal={IEEE Transactions on Neural Networks and Learning Systems}, 
  title={The Heidelberg Spiking Data Sets for the Systematic Evaluation of Spiking Neural Networks}, 
  year={2022},
  volume={33},
  number={7},
  pages={2744-2757},
  doi={10.1109/TNNLS.2020.3044364},
url = {https://zenkelab.org/resources/spiking-heidelberg-datasets-shd/}
}

@INPROCEEDINGS{yao2021,
  author={Yao, Man and Gao, Huanhuan and Zhao, Guangshe and Wang, Dingheng and Lin, Yihan and Yang, Zhaoxu and Li, Guoqi},
  booktitle={2021 IEEE/CVF International Conference on Computer Vision (ICCV)}, 
  title={Temporal-wise Attention Spiking Neural Networks for Event Streams Classification}, 
  year={2021},
  volume={},
  number={},
  pages={10201-10210},
  doi={10.1109/ICCV48922.2021.01006}}

@ARTICLE{yu2022,
AUTHOR={Yu, Chengting  and Gu, Zheming  and Li, Da  and Wang, Gaoang  and Wang, Aili  and Li, Erping },
TITLE={STSC-SNN: Spatio-Temporal Synaptic Connection with temporal convolution and attention for spiking neural networks},
JOURNAL={Frontiers in Neuroscience},
VOLUME={16},
YEAR={2022},
DOI={10.3389/fnins.2022.1079357},
ISSN={1662-453X}}

@InProceedings{dampfhoffer2022,
author="Dampfhoffer, Manon
and Mesquida, Thomas
and Valentian, Alexandre
and Anghel, Lorena",
editor="Pimenidis, Elias
and Angelov, Plamen
and Jayne, Chrisina
and Papaleonidas, Antonios
and Aydin, Mehmet",
title="Investigating Current-Based and Gating Approaches for Accurate and Energy-Efficient Spiking Recurrent Neural Networks",
booktitle="Artificial Neural Networks and Machine Learning -- ICANN 2022",
year="2022",
publisher="Springer Nature Switzerland",
address="Cham",
pages="359--370",
isbn="978-3-031-15934-3",
doi = {10.1007/978-3-031-15934-3_30}
}

@ARTICLE{bittar2022,
AUTHOR={Bittar, Alexandre  and Garner, Philip N. },
TITLE={A surrogate gradient spiking baseline for speech command recognition},
JOURNAL={Frontiers in Neuroscience},
VOLUME={16},
YEAR={2022},
DOI={10.3389/fnins.2022.865897},
ISSN={1662-453X}}

@article{rossbroich2022,
   title={Fluctuation-driven initialization for spiking neural network training},
   volume={2},
   ISSN={2634-4386},
   DOI={10.1088/2634-4386/ac97bb},
   number={4},
   journal={Neuromorphic Computing and Engineering},
   publisher={IOP Publishing},
   author={Rossbroich, Julian and Gygax, Julia and Zenke, Friedemann},
   year={2022},
   month=dec, pages={044016} }

@misc{hammouamri2023,
      title={Learning Delays in Spiking Neural Networks using Dilated Convolutions with Learnable Spacings}, 
      author={Ilyass Hammouamri and Ismail Khalfaoui-Hassani and Timothée Masquelier},
      year={2023},
      eprint={2306.17670},
      archivePrefix={arXiv},
      primaryClass={cs.NE},
      url={https://arxiv.org/abs/2306.17670}, 
}

@Article{perez2021,
author={Perez-Nieves, Nicolas
and Leung, Vincent C. H.
and Dragotti, Pier Luigi
and Goodman, Dan F. M.},
title={Neural heterogeneity promotes robust learning},
journal={Nature Communications},
year={2021},
month={Oct},
day={04},
volume={12},
number={1},
pages={5791},
issn={2041-1723},
doi={10.1038/s41467-021-26022-3},
url={https://doi.org/10.1038/s41467-021-26022-3}
}

@ARTICLE{sun2023,
    AUTHOR={Sun, Pengfei  and Chua, Yansong  and Devos, Paul  and Botteldooren, Dick },
    TITLE={Learnable axonal delay in spiking neural networks improves spoken word recognition},
    JOURNAL={Frontiers in Neuroscience},
    VOLUME={17},
    YEAR={2023},
    DOI={10.3389/fnins.2023.1275944},
    ISSN={1662-453X}
}

@article{dagostino2024,
      title={DenRAM: Neuromorphic Dendritic Architecture with RRAM for Efficient Temporal Processing with Delays}, 
      author={Simone D'Agostino and Filippo Moro and Tristan Torchet and Yigit Demirag and Laurent Grenouillet and Giacomo Indiveri and Elisa Vianello and Melika Payvand},
      JOURNAL={Nature Communications},
      year={2024},
      VOLUME={15},
      number={3446},
      url={https://doi.org/10.1038/s41467-024-47764-w}, 
}

@misc{malettira2024,
      title={{TSkips: Efficiency Through Explicit Temporal Delay Connections in Spiking Neural Networks}}, 
      author={Prajna G. Malettira and Shubham Negi and Wachirawit Ponghiran and Kaushik Roy},
      year={2024},
      eprint={2411.16711},
      archivePrefix={arXiv},
      primaryClass={cs.NE},
      url={https://arxiv.org/abs/2411.16711}, 
}

@INPROCEEDINGS{lars,
  author={Rafeldt, Lars and Mesquida, Thomas and Nakano, Hiroshi and Dampfhoffer, Manon and Moro, Filippo and Vivet, Pascal and Payvand, Melika and Dalgaty, Thomas},
  booktitle={2025 IEEE International Symposium on Circuits and Systems (ISCAS)}, 
  title={Event-based Audio Prediction with Spectro-Temporal Event-Graphs}, 
  year={2025},
  volume={},
  number={},
  pages={1-5},
  doi={10.1109/ISCAS56072.2025.11043865}}

@inproceedings{mostafa2024,
  title={17.8 0.4 V 988nW Time-Domain Audio Feature Extraction for Keyword Spotting Using Injection-Locked Oscillators},
  author={Mostafa, Ali and Hardy, Emmanuel and Badets, Franck},
  booktitle={2024 IEEE International Solid-State Circuits Conference (ISSCC)},
  volume={67},
  pages={328--330},
  year={2024},
  organization={IEEE},
  doi = {10.1109/ISSCC49657.2024.10454389}
}

@ARTICLE{dampfhoffer_tetci_2022,
  author={Dampfhoffer, Manon and Mesquida, Thomas and Valentian, Alexandre and Anghel, Lorena},
  journal={IEEE Transactions on Emerging Topics in Computational Intelligence}, 
  title={{Are SNNs Really More Energy-Efficient Than ANNs? an In-Depth Hardware-Aware Study}}, 
  year={2023},
  volume={7},
  number={3},
  pages={731-741},
  doi={10.1109/TETCI.2022.3214509}}

@misc{quantisenc2024,
      title={A Fully-Configurable Open-Source Software-Defined Digital Quantized Spiking Neural Core Architecture}, 
      author={Shadi Matinizadeh and Noah Pacik-Nelson and Ioannis Polykretis and Krupa Tishbi and Suman Kumar and M. L. Varshika and Arghavan Mohammadhassani and Abhishek Mishra and Nagarajan Kandasamy and James Shackleford and Eric Gallo and Anup Das},
      year={2024},
      eprint={2404.02248},
      archivePrefix={arXiv},
      primaryClass={cs.AR},
      url={https://arxiv.org/abs/2404.02248}, 
}

@inproceedings{jacob2018quantization,
  title={Quantization and training of neural networks for efficient integer-arithmetic-only inference},
  author={Jacob, Benoit and Kligys, Skirmantas and Chen, Bo and Zhu, Menglong and Tang, Matthew and Howard, Andrew and Adam, Hartwig and Kalenichenko, Dmitry},
  booktitle={Proceedings of the IEEE conference on computer vision and pattern recognition},
  pages={2704--2713},
  year={2018},
  doi = {10.1109/CVPR.2018.00286}
}

@INPROCEEDINGS{Al-Ameri,
  author={Al-Ameri, Yasir and Nguyen, Ming and Westerlund, Tomi},
  booktitle={2024 IEEE Nordic Circuits and Systems Conference (NorCAS)}, 
  title={FPGA-Based Hardware Acceleration for Deep Learning in Mobile Robotics}, 
  year={2024},
  volume={},
  number={},
  pages={1-7},
  doi={10.1109/NorCAS64408.2024.10752450}}

@INPROCEEDINGS{Al-Ali,
  author={Al-Ali, Firas and Gamage, Thilina Doremure and Nanayakkara, Hewa WTS and Mehdipour, Farhad and Ray, Sayan Kumar},
  booktitle={2020 IEEE Canadian Conference on Electrical and Computer Engineering (CCECE)}, 
  title={Novel Casestudy and Benchmarking of AlexNet for Edge AI: From CPU and GPU to FPGA}, 
  year={2020},
  volume={},
  number={},
  pages={1-4},
  doi={10.1109/CCECE47787.2020.9255739}}

@article{Guo,
author = {Guo, Kaiyuan and Zeng, Shulin and Yu, Jincheng and Wang, Yu and Yang, Huazhong},
title = {[DL] A Survey of FPGA-based Neural Network Inference Accelerators},
year = {2019},
issue_date = {March 2019},
publisher = {Association for Computing Machinery},
address = {New York, NY, USA},
volume = {12},
number = {1},
issn = {1936-7406},
doi = {10.1145/3289185},
journal = {ACM Trans. Reconfigurable Technol. Syst.},
month = mar,
articleno = {2},
numpages = {26},
}

@InProceedings{huber2024scaling,
author="Huber, Thomas E.
and Lecomte, Jules
and Polovnikov, Borislav
and von Arnim, Axel",
editor="Del Bue, Alessio
and Canton, Cristian
and Pont-Tuset, Jordi
and Tommasi, Tatiana",
title="Scaling Up Resonate-and-Fire Networks for Fast Deep Learning",
booktitle="Computer Vision -- ECCV 2024 Workshops",
year="2025",
publisher="Springer Nature Switzerland",
address="Cham",
pages="241--258",
isbn="978-3-031-92460-6",
doi="10.1007/978-3-031-92460-6_15"
}

@Article{Yin2021,
author={Yin, Bojian
and Corradi, Federico
and Boht{\'e}, Sander M.},
title={Accurate and efficient time-domain classification with adaptive spiking recurrent neural networks},
journal={Nature Machine Intelligence},
year={2021},
month={Oct},
day={01},
volume={3},
number={10},
pages={905-913},
issn={2522-5839},
doi={10.1038/s42256-021-00397-w},
url={https://doi.org/10.1038/s42256-021-00397-w}
}

@INPROCEEDINGS{Sadovsky2023,
  author={Sadovsky, Erik and Jakubec, Maros and Jarina, Roman},
  booktitle={2023 33rd International Conference Radioelektronika (RADIOELEKTRONIKA)}, 
  title={Speech Command Recognition Based on Convolutional Spiking Neural Networks}, 
  year={2023},
  volume={},
  number={},
  pages={1-5},
  doi={10.1109/RADIOELEKTRONIKA57919.2023.10109082}}

@article{nowotny2025loss,
doi = {10.1088/2634-4386/ada852},
url = {https://dx.doi.org/10.1088/2634-4386/ada852},
year = {2025},
month = {jan},
publisher = {IOP Publishing},
volume = {5},
number = {1},
pages = {014001},
author = {Nowotny, Thomas and Turner, James P and Knight, James C},
title = {Loss shaping enhances exact gradient learning with Eventprop in spiking neural networks},
journal = {Neuromorphic Computing and Engineering}
}

@article{ahlawat2025survey,
title = {Automatic Speech Recognition: A survey of deep learning techniques and approaches},
journal = {International Journal of Cognitive Computing in Engineering},
volume = {6},
pages = {201-237},
year = {2025},
issn = {2666-3074},
doi = {https://doi.org/10.1016/j.ijcce.2024.12.007},
url = {https://www.sciencedirect.com/science/article/pii/S2666307424000573},
author = {Harsh Ahlawat and Naveen Aggarwal and Deepti Gupta},
}

@ARTICLE{lopez2021kwsoverview,
  author={López-Espejo, Iván and Tan, Zheng-Hua and Hansen, John H. L. and Jensen, Jesper},
  journal={IEEE Access}, 
  title={Deep Spoken Keyword Spotting: An Overview}, 
  year={2022},
  volume={10},
  number={},
  pages={4169-4199},
  doi={10.1109/ACCESS.2021.3139508}}

@misc{warden2018googledataset,
      title={Speech Commands: A Dataset for Limited-Vocabulary Speech Recognition}, 
      author={Pete Warden},
      year={2018},
      eprint={1804.03209},
      archivePrefix={arXiv},
      primaryClass={cs.CL},
      url={https://arxiv.org/abs/1804.03209}, 
}

@InProceedings{nakano2025arc,
author="Nakano, Hiroshi
and Blachut, Krzysztof
and Jeziorek, Kamil
and Wzorek, Piotr
and Dampfhoffer, Manon
and Mesquida, Thomas
and Nishi, Hiroaki
and Kryjak, Tomasz
and Dalgaty, Thomas",
editor="Giorgi, Roberto
and Stojilovi{\'{c}}, Mirjana
and Stroobandt, Dirk
and Brox Jim{\'e}nez, Piedad
and Barriga Barros, {\'A}ngel",
title="Hardware-Accelerated Event-Graph Neural Networks for Low-Latency Time-Series Classification on SoC FPGA",
booktitle="Applied Reconfigurable Computing. Architectures, Tools, and Applications",
year="2025",
publisher="Springer Nature Switzerland",
address="Cham",
pages="51--68",
isbn="978-3-031-87995-1",
doi="10.1007/978-3-031-87995-1_4"
}

@INPROCEEDINGS{yoon2023,
  author={Yoon, Jinsung and Lee, Donghyun and Kim, Neungyun and Lee, Su-Jung and Kwak, Gil-Ho and Kim, Tae-Hwan},
  booktitle={2023 IEEE Symposium in Low-Power and High-Speed Chips (COOL CHIPS)}, 
  title={A Real-Time Keyword Spotting System Based on an End-To-End Binary Convolutional Neural Network in FPGA}, 
  year={2023},
  volume={},
  number={},
  pages={1-3},
  doi={10.1109/COOLCHIPS57690.2023.10121981}}

@INPROCEEDINGS{mourrane2023,
  author={Mourrane, Soufiane and Larras, Benoit and Clerc, Sylvain and Cathelin, Andreia and Frappé, Antoine},
  booktitle={2023 21st IEEE Interregional NEWCAS Conference (NEWCAS)}, 
  title={Low-Power Event-Driven Spectrogram Extractor for Multiple Keyword Spotting: A proof of concept}, 
  year={2023},
  volume={},
  number={},
  pages={1-5},
  doi={10.1109/NEWCAS57931.2023.10198120}}

@INPROCEEDINGS{krishna2023,
  author={Krishna, Adithya and Shankaranarayanan, H and Oleti, Hitesh Pavan and Chauhan, Anand and van Schaik, André and Mehendale, Mahesh and Thakur, Chetan Singh},
  booktitle={2023 IEEE Asia Pacific Conference On Postgraduate Research In Microelectronics And Electronics (PRIMEAsia)}, 
  title={TinyML Acoustic Classification using RAMAN Accelerator and Neuromorphic Cochlea}, 
  year={2023},
  volume={},
  number={},
  pages={44-45},
  doi={10.1109/PRIMEAsia60757.2023.00022}}

@INPROCEEDINGS{ng2024,
  author={Ng, Wei Soon and Ling Goh, Wang and Gao, Yuan},
  booktitle={2024 IEEE International Symposium on Circuits and Systems (ISCAS)}, 
  title={High Accuracy and Low Latency Mixed Precision Neural Network Acceleration for TinyML Applications on Resource-Constrained FPGAs}, 
  year={2024},
  volume={},
  number={},
  pages={1-5},
  doi={10.1109/ISCAS58744.2024.10558440}}

@INPROCEEDINGS{rohlicek1989,
  author={Rohlicek, J.R. and Russell, W. and Roukos, S. and Gish, H.},
  booktitle={International Conference on Acoustics, Speech, and Signal Processing,}, 
  title={Continuous hidden Markov modeling for speaker-independent word spotting}, 
  year={1989},
  volume={},
  number={},
  pages={627-630 vol.1},
  doi={10.1109/ICASSP.1989.266505}}

@INPROCEEDINGS{chen2014,
  author={Chen, Guoguo and Parada, Carolina and Heigold, Georg},
  booktitle={2014 IEEE International Conference on Acoustics, Speech and Signal Processing (ICASSP)}, 
  title={Small-footprint keyword spotting using deep neural networks}, 
  year={2014},
  volume={},
  number={},
  pages={4087-4091},
  doi={10.1109/ICASSP.2014.6854370}}

@ARTICLE{davis1980,
  author={Davis, S. and Mermelstein, P.},
  journal={IEEE Transactions on Acoustics, Speech, and Signal Processing}, 
  title={Comparison of parametric representations for monosyllabic word recognition in continuously spoken sentences}, 
  year={1980},
  volume={28},
  number={4},
  pages={357-366},
  doi={10.1109/TASSP.1980.1163420}}

@INPROCEEDINGS{ceolini2019,
  author={Ceolini, Enea and Anumula, Jithendar and Braun, Stefan and Liu, Shih-Chii},
  booktitle={ICASSP 2019 - 2019 IEEE International Conference on Acoustics, Speech and Signal Processing (ICASSP)}, 
  title={Event-driven Pipeline for Low-latency Low-compute Keyword Spotting and Speaker Verification System}, 
  year={2019},
  volume={},
  number={},
  pages={7953-7957},
  doi={10.1109/ICASSP.2019.8683669}}

@INPROCEEDINGS{blouw2020,
  author={Blouw, Peter and Eliasmith, Chris},
  booktitle={ICASSP 2020 - 2020 IEEE International Conference on Acoustics, Speech and Signal Processing (ICASSP)}, 
  title={Event-Driven Signal Processing with Neuromorphic Computing Systems}, 
  year={2020},
  volume={},
  number={},
  pages={8534-8538},
  doi={10.1109/ICASSP40776.2020.9053043}}

@INPROCEEDINGS{pedroni2018,
  author={Pedroni, Bruno U. and Sheik, Sadique and Mostafa, Hesham and Paul, Somnath and Augustine, Charles and Cauwenberghs, Gert},
  booktitle={2018 IEEE Biomedical Circuits and Systems Conference (BioCAS)}, 
  title={Small-footprint Spiking Neural Networks for Power-efficient Keyword Spotting}, 
  year={2018},
  volume={},
  number={},
  pages={1-4},
  doi={10.1109/BIOCAS.2018.8584832}}

@INPROCEEDINGS{dominguez2018nas,
  author={Dominguez-Morales, Juan P. and Liu, Qian and James, Robert and Gutierrez-Galan, Daniel and Jimenez-Fernandez, Angel and Davidson, Simon and Furber, Steve},
  booktitle={2018 International Joint Conference on Neural Networks (IJCNN)}, 
  title={Deep Spiking Neural Network model for time-variant signals classification: a real-time speech recognition approach}, 
  year={2018},
  volume={},
  number={},
  pages={1-8},
  doi={10.1109/IJCNN.2018.8489381}}

@INPROCEEDINGS{schone2024,
  author={Schöne, Mark and Sushma, Neeraj Mohan and Zhuge, Jingyue and Mayr, Christian and Subramoney, Anand and Kappel, David},
  booktitle={2024 International Conference on Neuromorphic Systems (ICONS)}, 
  title={Scalable Event-by-Event Processing of Neuromorphic Sensory Signals with Deep State-Space Models}, 
  year={2024},
  volume={},
  number={},
  pages={124-131},
  doi={10.1109/ICONS62911.2024.00026}}

@Article{baronig2025,
author={Baronig, Maximilian
and Ferrand, Romain
and Sabathiel, Silvester
and Legenstein, Robert},
title={Advancing spatio-temporal processing through adaptation in spiking neural networks},
journal={Nature Communications},
year={2025},
month={Jul},
day={01},
volume={16},
number={1},
pages={5776},
issn={2041-1723},
doi={10.1038/s41467-025-60878-z},
url={https://doi.org/10.1038/s41467-025-60878-z}
}

@article{sun2025,
title = {Towards parameter-free attentional spiking neural networks},
journal = {Neural Networks},
volume = {185},
pages = {107154},
year = {2025},
issn = {0893-6080},
doi = {https://doi.org/10.1016/j.neunet.2025.107154},
url = {https://www.sciencedirect.com/science/article/pii/S0893608025000334},
author = {Pengfei Sun and Jibin Wu and Paul Devos and Dick Botteldooren},
}

@inproceedings{kim21l_interspeech,
  author={Byeonggeun Kim and Simyung Chang and Jinkyu Lee and Dooyong Sung},
  title={{Broadcasted Residual Learning for Efficient Keyword Spotting}},
  year=2021,
  booktitle={Proc. Interspeech 2021},
  pages={4538--4542},
  doi={10.21437/Interspeech.2021-383}
}

@ARTICLE{lstm,

  author={Hochreiter, Sepp and Schmidhuber, Jürgen},

  journal={Neural Computation}, 

  title={Long Short-Term Memory}, 

  year={1997},

  volume={9},

  number={8},

  pages={1735-1780},

  doi={10.1162/neco.1997.9.8.1735}}

@misc{gru,
      title={On the Properties of Neural Machine Translation: Encoder-Decoder Approaches}, 
      author={Kyunghyun Cho and Bart van Merrienboer and Dzmitry Bahdanau and Yoshua Bengio},
      year={2014},
      eprint={1409.1259},
      archivePrefix={arXiv},
      primaryClass={cs.CL}
}

@article{kingma2014adam,
  title={Adam: A method for stochastic optimization},
  author={Kingma, Diederik P},
  journal={arXiv preprint arXiv:1412.6980},
  year={2014}
}

@article{paszke2017automatic,
  title={Automatic differentiation in pytorch},
  author={Paszke, Adam and Gross, Sam and Chintala, Soumith and Chanan, Gregory and Yang, Edward and DeVito, Zachary and Lin, Zeming and Desmaison, Alban and Antiga, Luca and Lerer, Adam},
  year={2017}
}

@software{Falcon_PyTorch_Lightning_2019,
author = {Falcon, William and {The PyTorch Lightning team}},
doi = {10.5281/zenodo.3828935},
license = {Apache-2.0},
month = mar,
title = {{PyTorch Lightning}},
url = {https://github.com/Lightning-AI/lightning},
version = {1.4},
year = {2019}
}





\end{document}